\documentclass{article}
\usepackage{iftex}
\usepackage{microtype}
\usepackage{graphicx}
\usepackage{subcaption}
\usepackage{booktabs}
\usepackage{hyperref}

\usepackage[preprint]{icml2026}
\usepackage{etoolbox}
\patchcmd{\printAffiliationsAndNotice}
  {{\bf AUTHORERR: Missing \textbackslash{}icmlcorrespondingauthor.}}
  {}{}{\PackageError{geogroup}{Could not omit correspondence placeholder}{}}
\usepackage{amsmath,amssymb,mathtools,amsthm}
\ifXeTeX\usepackage{fontspec}\fi
\usepackage[capitalize,noabbrev]{cleveref}
\crefname{figure}{Fig.}{Figs.}
\Crefname{figure}{Fig.}{Figs.}
\usepackage{array,multirow,makecell,xcolor,colortbl,tabularx,placeins,enumitem}
\usepackage{tikz}
\newtheorem{proposition}{Proposition}
\newcommand{\ggtabsetup}{\centering\small\setlength{\tabcolsep}{3pt}\renewcommand{\arraystretch}{1.10}}
\newcommand{\ggdefault}[1]{#1}
\AtBeginEnvironment{table}{\hypersetup{linkcolor=black,citecolor=black,urlcolor=black}}
\AtBeginEnvironment{table*}{\hypersetup{linkcolor=black,citecolor=black,urlcolor=black}}
\newcommand{\method}{GeoGroup}
\newcommand{\geopair}{GeoPair}
\newcommand{\natoms}{k}
\newcommand{\nnz}{s}
\newcommand{\pj}[1]{\textrm{#1}}
\newcommand{\na}{--}
\newcommand{\dense}[1]{#1}
\newcommand{\sd}[2]{#1{\scriptsize$\pm$#2}}
\newcommand{\stk}[2]{\makecell[c]{#1\\{\tiny$\pm$#2}}}

\definecolor{ggprimary}{HTML}{8E2A2A}
\definecolor{ggdeep}{HTML}{661E1E}
\definecolor{ggtint}{HTML}{E6D0D0}
\definecolor{ggsecond}{HTML}{416678}
\definecolor{ggsecondtint}{HTML}{D4DEE3}
\definecolor{ggband}{HTML}{F0E6CF}
\definecolor{ggline}{HTML}{BAB7AE}
\definecolor{ggbase}{HTML}{E4E2DE}
\definecolor{ggmuted}{HTML}{52514E}

\newcommand{\tablecolors}{\setlength{\aboverulesep}{0pt}\setlength{\belowrulesep}{0pt}\renewcommand{\arraystretch}{1.16}}
\newcommand{\rowband}{}
\newcommand{\rowgeopair}{}
\newcommand{\rowours}{}
\newcommand{\rowoursalt}{}

\newcommand{\model}[1]{\textit{#1}}
\renewcommand{\dense}[1]{#1}
\renewcommand{\stk}[2]{\makecell[c]{#1\\[-1.5pt]{\tiny\color{black}$\pm$#2}}}
\renewcommand{\sd}[2]{#1{\scriptsize\color{black}$\pm$#2}}
\newcolumntype{Y}{>{\centering\arraybackslash\hyphenpenalty=10000\exhyphenpenalty=10000}X}

\usepackage{soul}
\newif\ifrevisionhighlight
\revisionhighlightfalse
\sethlcolor{yellow!45}
\DeclareRobustCommand{\rev}[1]{\ifrevisionhighlight\hl{#1}\else#1\fi}
\makeatletter
\DeclareRobustCommand{\revmath}[1]{{\ifrevisionhighlight\mathpalette\revision@math{#1}\else#1\fi}}
\newcommand{\revision@math}[2]{\begingroup\setlength{\fboxsep}{0pt}\colorbox{yellow!45}{$\m@th#1#2$}\endgroup}
\makeatother
\soulregister\citep7
\soulregister\citet7
\soulregister\cite7
\soulregister\ref7
\soulregister\eqref7
\soulregister\method7
\soulregister\geopair7
\soulregister\pj7
\expandafter\def\expandafter\UrlBreaks\expandafter{\UrlBreaks\do\-}
\icmltitlerunning{GeoGroup: Geometry-Preserving Dictionary Sharing}
\begin{document}
\twocolumn[
\icmltitle{GeoGroup: Scalable Geometry-Preserving Dictionary Sharing for LLM Compression}
\begin{icmlauthorlist}
\icmlauthor{Nikolai Kaliazin}{itmo,mbzuai}
\icmlauthor{Magauiya Zhussip}{itmo,mws}
\icmlauthor{Baher Mohammad}{itmo,mws}
\icmlauthor{Ammar Ali}{itmo,mws}
\icmlauthor{Stamatios Lefkimmiatis}{itmo,mws}
\end{icmlauthorlist}
\icmlaffiliation{mbzuai}{MBZUAI}
\icmlaffiliation{itmo}{ITMO}
\icmlaffiliation{mws}{MWS AI}
\icmlkeywords{model compression, sparse dictionary learning, parameter sharing}
\vskip 0.3in
]
\printAffiliationsAndNotice{This work was conducted during Nikolai Kaliazin's internship at MWS AI.\par}
\hypersetup{pdfauthor={Nikolai Kaliazin, Magauiya Zhussip, Baher Mohammad, Ammar Ali, Stamatios Lefkimmiatis},pdftitle={GeoGroup: Scalable Geometry-Preserving Dictionary Sharing for LLM Compression},pdfsubject={Preprint}}
\begin{abstract}
Cross-layer redundancy offers a powerful opportunity to compress large language models, but sharing a representation across projections must account for the different inputs they receive. Existing approaches either fit all members using pooled input statistics, obscuring these differences, or retain projection-specific statistics while restricting sharing to pairs. We introduce \mbox{\method{}}, a training-free framework that shares sparse dictionaries across groups of arbitrary size and preserves each projection's activation geometry through a matrix-free preconditioned conjugate-gradient solver. Broader sharing spreads the cost of a dictionary across more projections, freeing storage for richer sparse coefficients. \rev{Ablations} of \rev{grouping}, \rev{storage allocation}, and \rev{preconditioning identify} consecutive grouping \rev{with Jacobi} as an effective \rev{low-cost default}, \rev{while pooled preconditioning offers further} accuracy \rev{at higher} cost. At 40\% nominal compression of attention and MLP projections, consecutive groups of sixteen retain 95.5\% and 97.7\% of dense seven-task accuracy on Llama-3-8B and Qwen3.5-9B, improving on GeoPair, a state-of-the-art pairwise factorization method, by 6.11 and 10.01 percentage points. \rev{The gains extend to Llama-3.2-1B}, \rev{Phi-4}, \rev{and Gemma-3-4B}, \rev{where GeoGroup improves on GeoPair by 5.23}, \rev{6.84}, \rev{and 10.18 points at the same nominal compression}. \rev{Selective gate}/\rev{up compression of Qwen3.8-27B at CR 0.6 achieves an} \textit{\rev{Avg7}} \rev{of 74.04}\%, \rev{compared with 74.37}\% \rev{for the dense model}, \rev{and accelerates full-model decode by} $\revmath{1.13}\times$, \rev{demonstrating a practical quality}--\rev{speed trade-off} beyond \rev{the primary ablation models}.
\end{abstract}

\section{Introduction}
\label{sec:intro}

Large language models have become central to modern language and multimodal systems, but their scale makes deployment demanding: billions of parameters must be stored and accessed during inference~\citep{grattafiori2024llama3,gemma2025gemma3}. Reducing this footprint without repeating expensive model training is therefore an important goal. Post-training compression addresses it by exploiting redundancy in pretrained weights. Quantization reduces the precision of stored values, while matrix factorization replaces dense weights with compact representations~\citep{frantar2023gptq,wang2025svd}. Factorization is particularly suited to exploiting repeated structure, because several weight columns can reuse the same basis vectors. \vspace{-1mm}

Within a projection, low-rank factorization represents all weight columns through a smaller common basis. Activation-aware methods improve this approximation by using a small calibration dataset to measure errors in the projection's outputs, rather than in its weights alone~\citep{yuan2023asvd,wang2025svd}. Sparse dictionary learning adds flexibility: each column selects a few atoms from a learned dictionary, so different columns can occupy different subspaces~\citep{shopkhoev2025cospadi,makhov2026compot}. Both representations reduce redundancy within individual matrices. However, storing a separate basis or dictionary for every projection can still duplicate structure that recurs across the model.\vspace{-1mm}

Cross-layer sharing exploits this additional redundancy by storing one dictionary for several projections and retaining only their individual coefficients. Basis Sharing and grouped CoSpaDi demonstrate the potential of this approach~\citep{wang2025basis,shopkhoev2025cospadi}. Sharing is still more demanding than concatenating weight matrices: projections at different depths and in different blocks receive different inputs. An approximation error along a frequently activated direction can substantially change one projection's output while barely affecting another. We call this input-dependent weighting of reconstruction error the projection's \emph{activation geometry}. A dictionary shared across projections must accommodate these distinct geometries and their weights. \vspace{-1mm}

Existing methods address this \rev{problem} in two ways. Basis Sharing and grouped CoSpaDi use a shared activation metric: they pool calibration statistics and apply the same directional weighting to every member's reconstruction error. This simplifies the joint fit, but a direction that is important to one projection may matter little to another. Pooling replaces those distinct sensitivities with a common weighting. GeoPair instead keeps a separate metric for each projection and obtains an analytic dictionary update for pairs~\citep{geopair}. This leaves a consequential question open: \emph{can a single dictionary serve many projections while preserving the reconstruction requirements of every member?} The pairwise solution does not extend directly. With more members, the dictionary update becomes a multi-term matrix equation whose factors generally cannot be diagonalized in common bases. \vspace{-1mm}

We introduce \method{}, a training-free framework that shares a sparse dictionary across arbitrarily large compatible groups while retaining each projection's activation metric. Matrix-free preconditioned conjugate gradients (PCG) solve the coupled dictionary equation without constructing its enormous coefficient matrix. A common preconditioner family connects inexpensive Jacobi scaling, pooled rotations, and Pair-GEVD. The latter recovers GeoPair's analytic dictionary update for two members, while Jacobi makes larger-group optimization practical with low setup and memory costs. \vspace{-1mm}

Group selection determines both reconstruction compatibility and representation capacity. Spreading the dictionary cost across more output columns leaves more storage for its atoms and coefficients. Since projections differ in output width, groups with the same member count can receive different capacities at the same compression ratio. We study this interaction through random, hierarchical activation-aware, and consecutive grouping, together with capacity controls and width-balancing experiments. Lower within-group distance alone does not identify the best partitions: storage allocation also matters. These experiments establish consecutive grouping in model order as an effective strategy combining competitive accuracy with negligible group-selection cost. \vspace{-1mm}

At 40\% nominal compression of attention and MLP projections, consecutive groups of sixteen achieve 66.61 and 70.58 mean accuracy over seven zero-shot tasks on Llama-3-8B and Qwen3.5-9B, respectively. Relative to GeoPair, a state-of-the-art method for geometry-preserving pairwise compression~\citep{geopair}, GeoGroup gains 6.11 and 10.01 percentage points, respectively, while \rev{retaining} 95.5\% and 97.7\% of the corresponding dense-model averages. \rev{The gains extend to Llama-3.2-1B}, \rev{Phi-4}, \rev{and Gemma-3-4B}, \rev{improving over GeoPair by 5.23}, \rev{6.84}, \rev{and 10.18 points at CR 0.4}. \rev{Selective gate}/\rev{up compression of Qwen3.8-27B achieves an} \textit{\rev{Avg7}} \rev{of 74.04}\%, \rev{compared with 74.37}\% \rev{for the dense model}, \rev{and yields a} $\revmath{1.13}\times$ \rev{full-model decode speedup at CR 0.6}. \rev{These evaluations establish effectiveness} across \rev{model families}, \rev{with additional evidence from DeepSeek-MoE}, multimodal \rev{tasks}, and factor quantization. Our contributions are:
\vspace{-2mm}
\begin{itemize}
\item \textbf{Geometry-preserving sharing beyond pairs.} We formulate shared sparse factorization for arbitrary compatible groups, retaining each projection's activation metric, and solve the resulting multi-term dictionary equation with matrix-free PCG. \vspace{-1mm}
\item \textbf{A preconditioner family that recovers GeoPair.} We unify Jacobi, pooled, and Pair-GEVD preconditioning. For groups of two, Pair-GEVD recovers GeoPair's dictionary update in \rev{a} single PCG step in exact arithmetic, with matching coefficients and regularization. \rev{Preconditioner and iteration-budget ablations quantify the accuracy}, \rev{runtime}, \rev{and memory trade-offs that motivate Jacobi as the default}. \vspace{-1mm}
\item \textbf{An extended study of group composition and capacity.} Comparisons of random, hierarchical, and consecutive groups, with capacity and width-balancing controls, show how storage allocation interacts with similarity and establish consecutive grouping as an effective strategy without similarity search.
\end{itemize}
\vspace{-2mm}

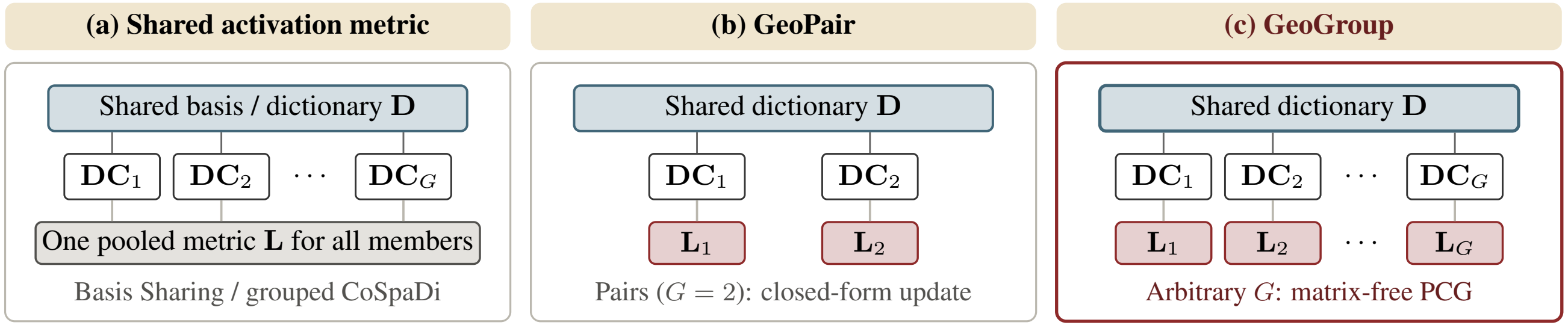
\begin{figure*}[t]
\centering
\begin{tikzpicture}[
  x=1cm, y=1cm, font=\fontsize{9}{10.5}\selectfont,
  card/.style={draw=ggline, line width=0.6pt, rounded corners=3pt},
  ours/.style={draw=ggprimary, line width=1pt, rounded corners=3pt},
  dict/.style={draw=ggsecond, fill=ggsecondtint, line width=0.75pt, rounded corners=2pt, minimum width=4.6cm, minimum height=0.5cm, inner sep=1.5pt},
  mem/.style={draw=black!80, fill=white, line width=0.55pt, rounded corners=1.5pt, minimum width=1.05cm, minimum height=0.5cm, inner sep=0pt},
  own/.style={draw=ggprimary, fill=ggtint, line width=0.65pt, rounded corners=2pt, minimum width=1.05cm, minimum height=0.46cm, inner sep=1pt},
  pool/.style={draw=ggmuted, fill=ggbase, line width=0.65pt, rounded corners=2pt, minimum width=4.6cm, minimum height=0.46cm, inner sep=2pt},
  lnk/.style={draw=black!60, line width=0.55pt},
  tie/.style={draw=ggline, line width=0.7pt},
  head/.style={font=\fontsize{9}{10.5}\selectfont\bfseries},
  note/.style={font=\fontsize{8.5}{10}\selectfont, text=ggmuted},
]
\draw[card] (0.000,-0.12) rectangle (5.550,2.72);
\draw[card] (5.775,-0.12) rectangle (11.325,2.72);
\draw[ours] (11.550,-0.12) rectangle (17.100,2.72);
\fill[ggband, rounded corners=3pt] (0.000,2.86) rectangle (5.550,3.38);
\fill[ggband, rounded corners=3pt] (5.775,2.86) rectangle (11.325,3.38);
\fill[ggband, rounded corners=3pt] (11.550,2.86) rectangle (17.100,3.38);
\node[head] at (2.775,3.12) {(a) Shared activation metric};
\node[head] at (8.550,3.12) {(b) \geopair{}};
\node[head, text=ggdeep] at (14.325,3.12) {(c) \method{}};
\node[dict] (Da) at (2.775,2.22) {Shared basis / dictionary $\mathbf{D}$};
\node[mem] (a1) at (1.175,1.47) {$\mathbf{D}\mathbf{C}_1$};
\node[mem] (a2) at (2.375,1.47) {$\mathbf{D}\mathbf{C}_2$};
\node at (3.375,1.47) {$\cdots$};
\node[mem] (aG) at (4.375,1.47) {$\mathbf{D}\mathbf{C}_G$};
\foreach \n in {a1,a2,aG} \draw[lnk] (Da.south -| \n.north) -- (\n.north);
\node[pool] (La) at (2.775,0.74) {One pooled metric $\mathbf{L}$ for all members};
\foreach \n in {a1,a2,aG} \draw[tie] (\n.south) -- (La.north -| \n.south);
\node[note] at (2.775,0.18) {Basis Sharing / grouped CoSpaDi};
\node[dict] (Db) at (8.550,2.22) {Shared dictionary $\mathbf{D}$};
\node[mem] (b1) at (7.600,1.47) {$\mathbf{D}\mathbf{C}_1$};
\node[mem] (b2) at (9.500,1.47) {$\mathbf{D}\mathbf{C}_2$};
\foreach \n in {b1,b2} \draw[lnk] (Db.south -| \n.north) -- (\n.north);
\node[own] (l1) at (7.600,0.74) {$\mathbf{L}_1$};
\node[own] (l2) at (9.500,0.74) {$\mathbf{L}_2$};
\draw[tie] (b1) -- (l1); \draw[tie] (b2) -- (l2);
\node[note] at (8.550,0.18) {Pairs ($G=2$): closed-form update};
\node[dict, line width=1pt] (Dc) at (14.325,2.22) {Shared dictionary $\mathbf{D}$};
\node[mem] (c1) at (12.725,1.47) {$\mathbf{D}\mathbf{C}_1$};
\node[mem] (c2) at (13.925,1.47) {$\mathbf{D}\mathbf{C}_2$};
\node at (14.925,1.47) {$\cdots$};
\node[mem] (cG) at (15.925,1.47) {$\mathbf{D}\mathbf{C}_G$};
\foreach \n in {c1,c2,cG} \draw[lnk] (Dc.south -| \n.north) -- (\n.north);
\node[own] (m1) at (12.725,0.74) {$\mathbf{L}_1$};
\node[own] (m2) at (13.925,0.74) {$\mathbf{L}_2$};
\node at (14.925,0.74) {$\cdots$};
\node[own] (mG) at (15.925,0.74) {$\mathbf{L}_G$};
\draw[tie] (c1) -- (m1); \draw[tie] (c2) -- (m2); \draw[tie] (cG) -- (mG);
\node[note, text=ggdeep] at (14.325,0.18) {Arbitrary $G$: matrix-free PCG};
\end{tikzpicture}
\caption{\textbf{Sharing a dictionary while preserving activation geometry.} Each $\mathbf{D}\mathbf{C}_i$ approximates $\mathbf{W}_i$. Blue marks the shared dictionary and red the individual activation metrics. (a) A pooled metric weights every member's error in the same way. (b) GeoPair retains individual metrics and obtains a closed-form dictionary update for pairs under its nonsingularity conditions. (c) GeoGroup retains individual metrics for arbitrary compatible groups and solves the coupled update with matrix-free PCG.}
\label{fig:overview}
\vspace{-2mm}
\end{figure*}

\section{Related Work}
\label{sec:related}

\paragraph{Post-training model compression.}
Post-training compression reduces the memory required to deploy a pretrained model without repeating its original training. Quantization lowers the number of bits used to store each parameter. GPTQ, for example, uses approximate second-order information to guide low-bit quantization~\citep{frantar2023gptq}. Factorization instead represents a weight matrix through smaller reusable factors. These approaches can be combined by quantizing the factors after they are fitted. Our focus is on sharing those factors across projections while preserving their distinct input-dependent behavior. \vspace{-3mm}

\paragraph{From low-rank approximation to sparse dictionaries.}
The quality of a compressed weight matrix depends on how its error propagates through the model. This motivates activation-aware factorization: ASVD incorporates activation statistics into the approximation, and SVD-LLM uses whitening to connect singular-value truncation to output reconstruction~\citep{yuan2023asvd,wang2025svd}. Dobi-SVD further learns how to allocate truncation across layers~\citep{wang2025dobi}. These methods express a matrix through a common low-dimensional subspace. Sparse dictionary learning relaxes this restriction by allowing each column to select its own subset of atoms. CoSpaDi adapts this representation to calibration-guided compression~\citep{shopkhoev2025cospadi}. \rev{Within this line of work}, ROCKET studies efficient optimization and compression allocation~\citep{ali2026rocket}, while COMPOT imposes dictionary orthogonality to simplify both dictionary and sparse coefficient updates~\citep{makhov2026compot}. \method{} addresses a different source of efficiency: reusing a dictionary across many projections while measuring each projection's reconstruction error using its own calibration inputs. \vspace{-3mm}

\paragraph{Cross-layer sharing and group selection.}
Basis Sharing demonstrates that projections in different layers can reuse a common basis, and grouped CoSpaDi extends reuse to sparse dictionaries~\citep{wang2025basis,shopkhoev2025cospadi}. Both fit the shared representation with a common activation metric derived from pooled calibration statistics. In this metric, an error in a given input direction receives the same weight for every projection, even when their own inputs excite that direction differently. GeoPair preserves separate reconstruction metrics and selects pairs through activation-aware graph matching, but its analytic dictionary update is specific to two members~\citep{geopair}. We extend this optimization to larger groups and examine whether increasingly broad sharing still benefits from similarity-based selection. MASA also shares atoms across attention layers, but learns matrix-valued atoms through gradient-based training~\citep{zhussip2026masa}. Our setting uses calibration-only factorization of a pretrained model. The distinction is thus both in what is shared and in how that shared representation is fitted. \vspace{-3mm}

\paragraph{Quantization of shared factors.}
Dictionary sharing and quantization reduce different parts of the storage cost: the former reduces the number of stored values, while the latter reduces their precision. We therefore also evaluate quantized GeoGroup factors. Sec.~\ref{sec:deployment} reports their accuracy and accounts for the storage of both coefficient values and sparse supports.
\vspace{-1mm}

\section{Geometry-Preserving Group Factorization}
\label{sec:method}

\method{} replaces a group of compatible projections with one shared dictionary and projection-specific sparse coefficients. The compression pipeline has three decisions: which projections share a dictionary, how much capacity their factors receive, and how those factors are fitted. We first define the reconstruction objective and the storage constraint, then describe group construction and the matrix-free solver that makes sharing beyond pairs practical.

\subsection{Shared Dictionaries with Individual Activation Geometry}
\label{sec:setup-overview}\label{sec:objective}
Consider a pretrained Transformer with projection weights $\mathbf{W}_i\in\mathbb R^{d\times m_i}$ and calibration inputs $\mathbf{X}_i\in\mathbb R^{N_i\times d}$. Its projection outputs are $\mathbf{X}_i\mathbf{W}_i$. For a group $\mathcal S$ of $G$ projections with the same input dimension $d$, we use
\begin{equation}
\mathbf{W}_i\approx \mathbf{D}\mathbf{C}_i,\quad \mathbf{D}\in\mathbb R^{d\times k},\quad \mathbf{C}_i\in\mathbb R^{k\times m_i}.
\label{eq:group-factorization}
\end{equation}
The columns of $\mathbf{D}$ are dictionary atoms, and each column of $\mathbf{C}_i$ has at most $s$ nonzero entries. Thus, each weight column uses a small subset of a common collection of atoms. Members can have different output widths $m_i$ and select different atoms. Only their input dimension must agree. The dictionary is stored once per group.

To preserve the behavior of projection $i$, we measure the output error $\|\mathbf{X}_i(\mathbf{W}_i-\mathbf{D}\mathbf{C}_i)\|_F^2$. Let $\mathbf{L}_i$ satisfy
$\mathbf{L}_i^\top \mathbf{L}_i=\mathbf{X}_i^\top \mathbf{X}_i+\eta_i \mathbf{I}$, with $\eta_i\ge0$ and a positive ridge when needed for nonsingularity. Then
\begin{equation}
\begin{aligned}
\|\mathbf{L}_i(\mathbf{W}_i-\mathbf{D}\mathbf{C}_i)\|_F^2
={}&\|\mathbf{X}_i(\mathbf{W}_i-\mathbf{D}\mathbf{C}_i)\|_F^2\\
&+\eta_i\|\mathbf{W}_i-\mathbf{D}\mathbf{C}_i\|_F^2.
\end{aligned}
\label{eq:main-activation-error}
\end{equation}
The factor $\mathbf{L}_i$ summarizes how the calibration inputs weight errors in different directions. Without the ridge, this is exactly output reconstruction error, and the ridge adds a weight-space penalty. We refer to this metric as the projection's activation geometry.

A natural way to fit shared factors is to concatenate the weights and use one whitening factor obtained from pooled activation statistics, as in Basis Sharing and grouped CoSpaDi~\citep{wang2025basis,shopkhoev2025cospadi}. This is a \emph{shared activation metric}: the same factor $\mathbf{L}$ measures every member's error as $\|\mathbf{L}(\mathbf{W}_i-\mathbf{D}\mathbf{C}_i)\|_F^2$, so errors along a given input direction receive the same weight across projections. It simplifies optimization, but replaces their individual directional weightings. If a direction is strongly activated in one projection and weakly activated in another, a pooled metric cannot represent both sensitivities. \method{} instead retains a separate $\mathbf{L}_i$ for each member. Defining $\alpha_i=\|\mathbf{L}_i\mathbf{W}_i\|_F^{-2}$ for nonzero output energies, we minimize
\begin{equation}
\begin{aligned}
\min_{\mathbf{D},\{\mathbf{C}_i\}}\quad &\mathcal L=\sum_{i\in\mathcal S}
\alpha_i\|\mathbf{L}_i(\mathbf{W}_i-\mathbf{D}\mathbf{C}_i)\|_F^2,\\
\text{subject to}\quad &\|\mathbf{C}_i[:,j]\|_0\le s\quad\text{for all }i,j.
\end{aligned}
\label{eq:objective}
\end{equation}
Normalizing by each member's energy prevents output scale alone from determining its contribution. The dictionary couples the fits, while the individual metrics and coefficients let members use it differently (Fig.~\ref{fig:overview}). For $G=1$, this is single-projection sparse factorization. For $G=2$, it is the relative-error version of the GeoPair objective. The normalizing weights can be absorbed into $\mathbf{L}_i$. Larger $G$ extends the same formulation to multiple projections.

\subsection{Turning Dictionary Sharing into Representation Capacity}
\label{sec:budget}
Sharing saves storage because each group stores its dictionary only once. We use these savings to increase the capacity of its factors at a fixed compression budget. Write $M=\sum_{i\in\mathcal S}m_i$ for the total output width. The original group contains $dM$ weight values, and its factors contain $dk$ dictionary values and at most $sM$ nonzero coefficient values. If $\mathrm{CR}$ is the fraction of values removed, the retained-value budget is
\begin{equation}
1-\mathrm{CR}=\frac{dk+sM}{dM}=\frac{k}{M}+\frac{s}{d}.
\label{eq:budget}
\end{equation}
The two terms expose the trade-off: $k/M$ is the dictionary cost per original weight, and $s/d$ is the coefficient cost. Increasing the group's width lowers the first cost for a fixed dictionary, leaving more room for coefficients.

To determine both $k$ and $s$, we start from a dictionary-to-support ratio $\kappa=k/s=2.5$. Solving Eq.~\eqref{eq:budget}, capping the dictionary at $\min(d,M)$, and assigning the remaining budget to coefficients gives
\begin{equation}
\begin{aligned}
k&=\min\left(\frac{(1-\mathrm{CR})dM}{d+M/\kappa},d,M\right),\\
s&=\frac{(1-\mathrm{CR})dM-dk}{M}.
\end{aligned}
\label{eq:support-budget}
\end{equation}
These expressions describe the allocation before integer rounding. Before the cap is reached, a wider group can afford a wider dictionary. Once $k=d$, its dictionary costs $d^2$ values, leaving $s=(1-\mathrm{CR})d-d^2/M$ nonzeros per output column.
For example, at CR $=0.4$, groups of one, two, and sixteen square projections allow approximately $0.17d$, $0.27d$, and $0.54d$ nonzeros per column, respectively. Sharing therefore enables richer representations even though the retained fraction stays at 60\%.

This benefit depends on total width $M$, not just member count $G$. Two groups of the same size can receive different capacities if their projections have different output widths. Group selection must therefore balance reconstruction compatibility against dictionary overhead. Throughout the paper, nominal CR counts values in the targeted projections. Sparse-position metadata and untouched model components are included separately in the storage accounting of Sec.~\ref{sec:deployment}.

\subsection{From Random Partitions to Consecutive Groups}
\label{sec:grouping}
We first partition the model's projections into pools with equal input dimension. Within each pool, the target group size $G$ controls how broadly a dictionary is shared. We compare three ways to choose its members, using the same objective and allocation rule for all three.\vspace{-2mm}

\paragraph{Random grouping: testing the effect of sharing.}
A straightforward approach is to shuffle a pool uniformly at random and divide it into groups of $G$ projections. This requires no assumptions about which layers have similar weights or activations. It provides a useful baseline: if larger random groups improve accuracy, their benefit cannot depend entirely on discovering a carefully selected partition. We repeat this procedure over multiple seeds to measure sensitivity to membership.\vspace{-3mm}

\paragraph{Hierarchical Blossom: selecting similar projections.}
A more informed approach groups projections whose weights are similar under their activation metrics. We build on GeoPair's graph-matching strategy, which reports state-of-the-art compression results among its evaluated factorization methods~\citep{geopair}. For $m_i<m_j$, GeoPair compares the narrower projection with the best contiguous column window of the wider projection, whitening each with its own calibration factor:
\begin{equation}
\delta_{ij}=\min_{0\le t\le m_j-m_i}
\frac{\|\mathbf{L}_j\mathbf{W}_j[:,t:t+m_i]-\mathbf{L}_i\mathbf{W}_i\|_F}
{\|\mathbf{L}_i\mathbf{W}_i\|_F+\epsilon}.
\label{eq:main-group-distance}
\end{equation}
Here $\epsilon>0$ stabilizes the normalization, and $\mathbf{W}_j[:,t:t+m_i]$ is a contiguous window of $m_i$ columns. This permits comparisons across projection types with different output widths. Equal-width comparisons require no window search. We treat projections as graph vertices and these discrepancies as edge costs. Minimum-cost maximum-cardinality matching, solved by Edmonds' Blossom algorithm~\citep{edmonds1965paths}, chooses disjoint pairs with minimum total cost among matchings of maximum size. Unlike choosing each projection's nearest neighbor independently, matching resolves competing choices jointly.

To extend pairing to $G>2$, \emph{Hierarchical Blossom} (HB) treats each newly formed group as a vertex at the next level. The cost between groups is their mean cross-member distance,
\begin{equation}
\delta(\mathcal S_a,\mathcal S_b)=
\frac{\sum_{i\in\mathcal S_a}\sum_{j\in\mathcal S_b}\delta_{ij}}
{|\mathcal S_a|\,|\mathcal S_b|}.
\label{eq:main-merge-distance}
\end{equation}
Repeated matching merges pairs into groups of four, then eight, and so on up to the tested target size. Each matching stage is optimal for its current edge costs, but earlier merges are fixed, so the hierarchy does not guarantee a globally optimal larger-group partition or maximum downstream accuracy. It also requires pairwise distance computation and repeated matching. \vspace{-3mm}

\paragraph{Consecutive grouping: effective sharing without search.}
The third approach uses the model's existing order. We sort projections by Transformer layer and a fixed within-layer projection order, then take consecutive groups of $G$ within each compatible pool. Our default \emph{Mixed} scheme allows attention and MLP projections to share a dictionary. In Llama-3-8B, for example, query, key, value, output, gate, and up projections form one input-dimension pool, while down projections form another. This mixes narrow and wide outputs across nearby layers without computing any similarity scores. Unless stated otherwise, leftover projections that cannot fill a target-size group are processed as singletons.

Our grouping study establishes Consecutive as the default for its favorable balance of compression quality and group-selection cost (Sec.~\ref{sec:exp-grouping}). Appendix~\ref{app:grouping-all} examines ordering, projection-type restrictions, and width-balanced alternatives.

\subsection{Solving the Multi-Projection Dictionary Problem}
\label{sec:alternating}\label{sec:solver}
Once groups and factor sizes are fixed, we alternate sparse coefficient updates with a joint dictionary update. The coefficient problems remain independent across projections. The shared dictionary is where their distinct activation geometries interact. Our main computational contribution is to solve this coupled update for arbitrary $G$ using matrix-free PCG with Jacobi preconditioning. \vspace{-3mm}

\paragraph{Initialization and sparse coefficients.}
We initialize $\mathbf{D}$ with the leading $k$ left singular vectors of the concatenated unwhitened weights $[\mathbf{W}_i]_{i\in\mathcal S}$ and initialize coefficients by activation-weighted least squares. With $\mathbf{D}$ fixed, each weight column $\mathbf{w}$ requires a sparse code minimizing $\|\mathbf{L}_i(\mathbf{w}-\mathbf{D}\mathbf{c})\|_2^2$ subject to $\|\mathbf{c}\|_0\le s$. We perform two consecutive hard thresholding pursuit (HTP) steps~\citep{foucart2011htp} per coefficient update. Each step proposes coefficients using a gradient update, retains the $s$ largest-magnitude entries, and refits values on that support with 10 CG iterations. The subsequent dictionary step uses these sparse codes. \vspace{-3mm}

\paragraph{Why the dictionary update becomes difficult.}
For fixed coefficients, define
$\mathbf{A}_i=\alpha_i\mathbf{L}_i^\top \mathbf{L}_i$, $\mathbf{B}_i=\mathbf{C}_i\mathbf{C}_i^\top$, and $\mathbf{K}=\sum_i \mathbf{A}_i\mathbf{W}_i\mathbf{C}_i^\top$.
Setting the derivative of Eq.~\eqref{eq:objective} to zero gives
\begin{equation}
\mathcal A(\mathbf{D})\equiv\sum_{i\in\mathcal S}\mathbf{A}_i\mathbf{D}\mathbf{B}_i=\mathbf{K}.
\label{eq:sylvester}
\vspace{-1mm}
\end{equation}
Each member contributes a term coupling its activation metric $\mathbf{A}_i$ with its coefficient Gram matrix $\mathbf{B}_i$. This is a multi-term generalized Sylvester equation. The two-term case admits the GeoPair reduction in Eq.~\eqref{eq:main-pair-recovery}, but three or more terms generally cannot be diagonalized in the same pair of bases. A generic direct solve would vectorize $\mathbf{D}$ and construct the matrix $\sum_i \mathbf{B}_i\otimes \mathbf{A}_i$ of size $(dk)\times(dk)$. Even at $d=k=4096$, that matrix would contain approximately $2.8\times10^{14}$ entries. \vspace{-3mm}

\paragraph{Matrix-free PCG with Jacobi preconditioning.}
We avoid this construction by applying PCG~\citep{saad2003iterative} directly to dictionary-shaped matrices. It needs only products with the operator,
\begin{equation}
\mathcal A(\mathbf{Z})=\sum_{i\in\mathcal S}\alpha_i\mathbf{L}_i^\top(\mathbf{L}_i\mathbf{Z}\mathbf{C}_i)\mathbf{C}_i^\top,
\label{eq:matrix-free-operator}
\end{equation}
which can be evaluated with the individual factors. Every projection retains its own metric throughout the solve. The operator is self-adjoint and satisfies
$\langle \mathbf{Z},\mathcal A(\mathbf{Z})\rangle_F=\sum_i\alpha_i\|\mathbf{L}_i\mathbf{Z}\mathbf{C}_i\|_F^2\ge0$.
We use the activation ridge in Eq.~\eqref{eq:main-activation-error} to make $\mathbf{L}_i$ nonsingular when needed. If the $\mathbf{L}_i$ are nonsingular and the concatenated codes $[\mathbf{C}_i]$ have full row rank, it is positive definite, so standard PCG applies. Without that rank condition, the dictionary minimizer may be nonunique. The coefficient ridge $\epsilon=10^{-6}$ stabilizes the coefficient solves but does not enforce this rank condition (Appendix~\ref{app:optimization-derivations}).

Jacobi preconditioning rescales each dictionary coordinate by the corresponding diagonal entry of the operator. For a residual matrix $\mathbf{R}$, we use
\begin{equation}
\begin{aligned}
\Delta_{ab}&=\sum_{i\in\mathcal S}(\mathbf{A}_i)_{aa}(\mathbf{B}_i)_{bb},\\
[\mathcal M^{-1}(\mathbf{R})]_{ab}&=R_{ab}/\max(\Delta_{ab},10^{-8}).
\end{aligned}
\label{eq:main-jacobi}
\end{equation}
This requires only a $d\times k$ array and elementwise division, avoiding the eigendecompositions and basis transformations of stronger preconditioners. The floor stabilizes division but does not remove a nullspace of $\mathcal A$. We warm-start from the previous dictionary and stop at relative residual $\|\mathbf{K}-\mathcal A(\mathbf{D})\|_F/\|\mathbf{K}\|_F\le10^{-4}$ or 20 iterations. In Sec.~\ref{sec:exp-solver}, we evaluate the resulting trade-offs among accuracy, runtime, and memory. \vspace{-4mm}

\paragraph{A preconditioner family that recovers GeoPair.}
Our solver places pairwise and groupwise dictionary learning in one framework. Beyond Jacobi, we consider diagonal preconditioning in transformed bases $\mathbf{U},\mathbf{V}$: transform the residual to $\mathbf{U}^\top \mathbf{R}\mathbf{V}$, divide by the diagonal of the transformed operator, and map back with $\mathbf{U}$ and $\mathbf{V}^\top$ (Appendix~\ref{app:optimization-derivations}). Identity bases give Jacobi, eigenbases of the summed factors give the pooled variant, and generalized eigenbases of a reference pair give Pair-GEVD. All three preserve the individual activation metrics in the objective.

For $G=2$, this family contains the exact GeoPair dictionary update. To match its regularized formulation, write $\widetilde{\mathbf{B}}_i=\mathbf{C}_i\mathbf{C}_i^\top+\epsilon_D \mathbf{I}$ and solve $\mathbf{A}_1\mathbf{D}\widetilde{\mathbf{B}}_1+\mathbf{A}_2\mathbf{D}\widetilde{\mathbf{B}}_2=\mathbf{K}$. This adds the dictionary penalty $\epsilon_D\sum_i\alpha_i\|\mathbf{L}_i\mathbf{D}\|_F^2$ to Eq.~\eqref{eq:objective}. Setting $\epsilon_D=0$ recovers our unregularized equation. When $\mathbf{A}_i$ and $\widetilde{\mathbf{B}}_i$ are positive definite, their generalized eigenbases $\mathbf{P},\mathbf{Q}$ satisfy
$\mathbf{P}^\top \mathbf{A}_1\mathbf{P}=\mathbf{I}$, $\mathbf{P}^\top \mathbf{A}_2\mathbf{P}=\boldsymbol{\Lambda}$,
$\mathbf{Q}^\top\widetilde{\mathbf{B}}_1\mathbf{Q}=\mathbf{I}$, and $\mathbf{Q}^\top\widetilde{\mathbf{B}}_2\mathbf{Q}=\boldsymbol{\Theta}$.
Substituting $\mathbf{D}=\mathbf{P}\mathbf{Y}\mathbf{Q}^\top$ gives $\mathbf{Y}+\boldsymbol{\Lambda} \mathbf{Y}\boldsymbol{\Theta}=\mathbf{P}^\top \mathbf{K}\mathbf{Q}$, hence
\begin{equation}
\mathbf{D}=\mathbf{P}\left[(\mathbf{P}^\top \mathbf{K}\mathbf{Q})\oslash
(\mathbf 1_d\mathbf 1_k^\top+\boldsymbol{\lambda}\boldsymbol{\theta}^\top)\right]\mathbf{Q}^\top,
\label{eq:main-pair-recovery}
\end{equation}
where $\boldsymbol{\lambda},\boldsymbol{\theta}$ are the diagonals of $\boldsymbol{\Lambda},\boldsymbol{\Theta}$ and $\oslash$ denotes elementwise division. Pair-GEVD is therefore an exact inverse for this two-member operator: from the SVD initialization or any subsequent warm start, PCG reaches GeoPair's dictionary solution in at most one iteration in exact arithmetic, provided the coefficients and regularization match. This is one dictionary solve, not one complete alternating fit. For larger groups, same construction becomes a preconditioner for the multi-term equation. Jacobi provides our lower-cost default, extending geometry-preserving sharing to groups for which the pairwise closed form no longer applies.

Algorithm~\ref{alg:geogroup} summarizes the fit. We run up to 30 alternating rounds, stop after ten rounds without improvement, and retain the best feasible factors. We assess numerical behavior through extensive preconditioner and iteration-budget ablations (Sec.~\ref{sec:exp-solver} and Appendix~\ref{app:solver}), complemented by an analysis of coefficient regularization and the rank conditions that govern uniqueness (Appendix~\ref{app:optimization-derivations}).
\begin{algorithm}[t]
\caption{GeoGroup for a compatible group $\mathcal S$}
\label{alg:geogroup}
\begin{algorithmic}[1]
\REQUIRE $\{\mathbf{W}_i,\mathbf{L}_i\}_{i\in\mathcal S}$, nominal CR
\STATE Allocate $(k,s)$ by Eq.~\eqref{eq:support-budget}.
\STATE Initialize $\mathbf{D}$ by SVD of $[\mathbf{W}_i]$; solve initial codes.
\STATE $\ell^\star\gets\infty$; $t^\star\gets0$.
\FOR{$t=1,\ldots,30$}
\FORALL{$i\in\mathcal S$}
\STATE Update $\mathbf{C}_i$ by two HTP steps (10 CG per refit).
\ENDFOR
\STATE Solve Eq.~\eqref{eq:sylvester} by warm-started Jacobi-PCG
\STATE \hspace{1em}(at most 20 iterations; relative residual $10^{-4}$).
\IF{$\mathcal L(\mathbf{D},\{\mathbf{C}_i\})<\ell^\star$}
\STATE Save factors; $\ell^\star\gets\mathcal L$; $t^\star\gets t$.
\ENDIF
\IF{$t-t^\star\ge10$}
\STATE Stop.
\ENDIF
\ENDFOR
\STATE \textbf{return} saved factors.
\end{algorithmic}
\end{algorithm}
\vspace{-2mm}

\section{Experiments}
\label{sec:experiments}
Our experiments assess compression quality, investigate the roles of geometry, capacity, and grouping, and evaluate the dictionary solver.

\subsection{Experimental Setup}
\label{sec:exp-setup}
We \rev{evaluate GeoGroup across} Llama-3-8B, Qwen3.5-9B~\citep{grattafiori2024llama3,qwen2026qwen35}, Llama-3.2-1B~\citep{meta2024llama32}, Phi-4~\citep{abdin2024phi4}, Gemma-3-4B~\citep{gemma2025gemma3}, and Qwen3.8-27B. \rev{Tables}~\ref{tab:main-summary} and~\ref{tab:cross-model} \rev{summarize quality across these models}. \rev{Detailed ablations use Llama-3-8B and Qwen3.5-9B}. \rev{An additional DeepSeek-MoE study appears in Appendix}~\ref{app:moe}. We compress language-backbone attention and MLP projections, including Qwen's linear attention. Embeddings, output heads, and vision encoders remain dense. Unless specified otherwise, we use nominal CR $=0.4$, consecutive groups of $G=16$, and the optimization settings in Sec.~\ref{sec:alternating}. Factorization runs use NVIDIA A100 40GB GPUs: two for model-level compression and one for per-group solver diagnostics.

We evaluate seven zero-shot tasks using LM Evaluation Harness~\citep{eval-harness}: ARC-Challenge, ARC-Easy~\citep{clark2018arc}, HellaSwag~\citep{zellers2019hellaswag}, MMLU~\citep{hendrycks2021mmlu}, PIQA~\citep{bisk2020piqa}, RACE~\citep{lai2017race}, and SciQ~\citep{welbl2017sciq}. We report normalized accuracy for tasks that support this metric and standard accuracy for the remaining tasks. \textit{\rev{Avg7}} denotes the arithmetic mean of the seven task scores, computed before rounding. All text-only methods evaluated in our experiments use the same calibration set of 256 FineWeb sequences, each containing 1,024 tokens~\citep{penedo2024fineweb}. LAMBADA, WikiText-2, and C4 perplexity~\citep{paperno2016lambada,merity2017pointer,raffel2020exploring} complement the accuracy evaluation. Appendix~\ref{app:setup} details the protocol and normalization effects.

Our baselines include SVD-LLM without parameter updates, the official Basis Sharing implementation, and GeoPair with Blossom matching and its analytic dictionary update. GeoPair uses 50 rounds versus 30 for GeoGroup, and pair-PCG controls retain its Blossom pairs. \rev{We rerun} ROCKET and COMPOT \rev{under the same experimental conditions as GeoGroup and GeoPair}, \rev{using the common} calibration \rev{set}, \rev{compression scope}, and \rev{evaluation protocol}. Full task-level comparisons are provided in Appendix~\ref{app:full}.

\subsection{Effectiveness of Larger-Group Sharing}
\label{sec:exp-main}\label{sec:exp-transfer}
\begin{table}[t]
\caption{\textbf{Compression quality across models.} \rev{Average accuracy} (\%) and LAMBADA perplexity (PPL). Arrows indicate the preferred direction. Bold marks the best compressed result. \rev{ROCKET and COMPOT are rerun under the same experimental conditions as GeoGroup and GeoPair}. \rev{Detailed} results \rev{appear} in Table~\ref{tab:main}.}
\label{tab:main-summary}
\ggtabsetup\footnotesize
\setlength{\tabcolsep}{1pt}
\tablecolors
\begin{tabularx}{\columnwidth}{@{\hspace{1pt}}lc>{\hsize=.85\hsize\linewidth=\hsize}Y>{\hsize=1.15\hsize\linewidth=\hsize}Y>{\hsize=.85\hsize\linewidth=\hsize}Y>{\hsize=1.15\hsize\linewidth=\hsize}Y@{\hspace{1pt}}}
\toprule
\rowband & & \multicolumn{2}{c}{CR~$=0.4$} & \multicolumn{2}{c}{CR $=0.5$} \\
\cmidrule(lr){3-4}\cmidrule(l){5-6}
\rowband Method & $G$ & \textit{\rev{Avg7}}~$\uparrow$ & PPL~$\downarrow$ & \textit{\rev{Avg7}}~$\uparrow$ & \rev{PPL}~$\downarrow$ \\
\midrule
\model{Llama-3-8B}, dense & -- & 69.78 & 3.13 & 69.78 & \rev{3.13}\\
SVD-LLM & 1 & 33.82 & \mbox{$3.2\!\times\!10^3$} & 28.75 & \mbox{$\revmath{7.3}\!\times\!\revmath{10}^\revmath{4}$}\\
ROCKET & 1 & \rev{55.51} & \rev{10.04} & \rev{44.43} & \rev{164.6}\\
COMPOT & 1 & \rev{52.71} & \rev{21.35} & \rev{40.67} & \rev{530.6}\\
Basis Sharing & 2 & 35.04 & \mbox{$2.9\!\times\!10^3$} & 29.59 & \mbox{$\revmath{7.1}\!\times\!\revmath{10}^\revmath{4}$}\\
\rowgeopair GeoPair & 2 & 60.50 & 5.21 & 48.65 & \rev{23.42}\\
\rowours \ggdefault{GeoGroup (Consecutive)} & 16 & \textbf{66.61} & \textbf{2.78} & \textbf{61.09} & \textbf{\rev{3.83}}\\
\midrule
\model{Qwen3.5-9B}, dense & -- & 72.26 & 3.88 & 72.26 & \rev{3.88}\\
SVD-LLM & 1 & 41.67 & 65.36 & 35.51 & \rev{814.1}\\
\rev{ROCKET} & \rev{1} & \rev{60.27} & \rev{6.25} & \rev{48.87} & \rev{15.76}\\
\rev{COMPOT} & \rev{1} & \rev{59.17} & \rev{5.62} & \rev{47.53} & \rev{14.81}\\
Basis Sharing & 2 & 45.20 & 21.43 & 38.51 & \rev{86.95}\\
\rowgeopair GeoPair & 2 & 60.57 & 8.72 & 42.06 & \rev{153.4}\\
\rowours \ggdefault{GeoGroup (Consecutive)} & 16 & \textbf{70.58} & \textbf{3.31} & \textbf{64.38} & \textbf{\rev{4.47}}\\
\bottomrule
\end{tabularx}
\vspace{-2mm}
\end{table}

\begin{figure}[t]
\centering
\begin{tikzpicture}
\node[anchor=south west,inner sep=0] (plot) at (0,0)
  {\includegraphics[width=\columnwidth]{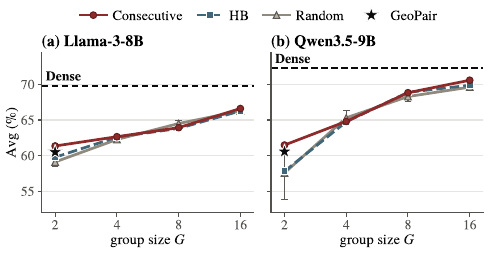}};
\begin{scope}[x={(plot.south east)},y={(plot.north west)}]
\fill[white] (0,0.36) rectangle (0.040,0.62);
\node[rotate=90,inner sep=0,font=\fontsize{7.5}{8}\selectfont]
  at (0.024,0.49) {\textit{Avg7} (\%)};
\end{scope}
\end{tikzpicture}
\caption{\textbf{Compression quality improves with broader sharing.} Nominal CR $0.4$, with common axes across models. Random error bars show one standard deviation over grouping seeds. HB pairs use Jacobi PCG, and the star marks the analytic GeoPair baseline. Detailed results appear in Table~\ref{tab:app-sweep-all}.}
\label{fig:group-size}
\vspace{-2mm}
\end{figure}

\paragraph{Quality across group sizes.}
Sharing beyond pairs substantially improves compression quality on \rev{Llama-3-8B and Qwen3.5-9B} (see Table~\ref{tab:main-summary}). At CR 0.4, consecutive $G=16$ achieves 66.61\% on Llama and 70.58\% on Qwen, exceeding GeoPair by 6.11 and 10.01 points and retaining 95.5\% and 97.7\% of the dense-model averages. The gains increase to 12.4 and 22.3 points at CR 0.5. This trend extends across grouping rules: at CR 0.4, the average accuracy increases with every doubling of $G$ from 2 to 16 for HB, Random, and Consecutive (see Fig.~\ref{fig:group-size}). Consecutive pairs achieve 61.36\% and 61.49\%, respectively, indicating the improvement with larger groups also holds within a fixed grouping strategy and dictionary solver.

\rev{On Llama-3-8B}, \rev{ROCKET and COMPOT achieve} \textit{\rev{Avg7}} \rev{scores of 55.51 and 52.71 at CR 0.4}, \rev{and 44.43 and 40.67 at CR 0.5}, \rev{respectively}. \rev{Consecutive} $\revmath{G}=\revmath{16}$ \rev{exceeds these by 11.10}/\rev{13.90 and 16.66}/\rev{20.42 points}, \rev{respectively}, \rev{and also has lower LAMBADA perplexity} (\rev{Table}~\ref{tab:main}).

Task-specific metrics reveal complementary effects. GeoGroup improves LAMBADA perplexity over GeoPair and the dense references, whereas WikiText-2 perplexity remains above dense: 9.07 versus 6.14 on Llama and 11.59 versus 8.65 on Qwen. Generation tasks are more sensitive: on Llama, GSM8K accuracy changes from 48.4 to 21.0 and HumanEval pass@1 from 37.8 to 7.3. Table~\ref{tab:main} gives LAMBADA, WikiText-2, and C4 perplexities and Appendix~\ref{app:bench} the generation tasks. \vspace{-3mm}

\begin{table}[t]
\caption{\textbf{\rev{Average accuracy} (\%) \rev{over seven zero-shot benchmarks across additional models}.} \rev{Bold marks the better compressed result}.}
\label{tab:cross-model}
\ggtabsetup
\setlength{\tabcolsep}{2pt}
\begin{tabularx}{\columnwidth}{@{}lYYYY@{}}
\toprule
\rev{Model} & \rev{CR} & \rev{Dense} & $\revmath{G}=\revmath{2}$ & $\revmath{G}=\revmath{16}$\\
\midrule
\rev{Llama-3.2-1B} & \rev{0.4} & \rev{57.71} & \rev{46.22} & \textbf{\rev{51.45}}\\
\rev{Phi-4} & \rev{0.4} & \rev{72.39} & \rev{65.23} & \textbf{\rev{72.07}}\\
\rev{Gemma-3-4B} & \rev{0.4} & \rev{56.95} & \rev{44.77} & \textbf{\rev{54.95}}\\
\rev{Qwen3.8-27B} & \rev{0.6} & \rev{74.37} & \rev{73.14} & \textbf{\rev{74.04}}\\
\bottomrule
\end{tabularx}
\end{table}

\paragraph{\rev{Effectiveness} across \rev{model families}.}
\rev{GeoGroup improves over GeoPair across all three additional full-backbone evaluations} (\rev{Table}~\ref{tab:cross-model}). \rev{At} CR 0.4, consecutive $G=16$ \rev{gains 5.23 points on Llama-3.2-1B}, \rev{6.84 on Phi-4}, \rev{and 10.18 on Gemma-3-4B}, \rev{reaching} 51.45, 72.07, and 54.95 \textit{\rev{Avg7}}. \rev{Phi-4 remains within 0.32 points of dense}, and \rev{Gemma within 2.00 points}. \rev{These results show that the benefit extends across model families and scales}, \rev{beyond the two models used for detailed ablations} (Appendix~\ref{app:transfer}). On multimodal Qwen3.5-9B, mixed text--image calibration yields 63.60 mean accuracy across six tasks, versus 57.33 for GeoPair and 68.48 dense. Appendix~\ref{app:mm} examines how including image inputs helps preserve visual capabilities. \vspace{-1mm}

\rev{On Qwen3.8-27B}, \rev{selective gate}/up compression \rev{at CR 0.6 retains 74.04} \textit{\rev{Avg7}} \rev{versus 74.37 dense and 73.14 for pairs} (\rev{Table}~\ref{tab:cross-model}). \rev{These same checkpoints provide} the \rev{quality}--\rev{speed comparison in Section}~\ref{sec:deployment}. \rev{Larger groups also improve} \textit{\rev{Avg7}} \rev{at CR 0.7}, \rev{while pairs perform better at CR 0.9}. \rev{LAMBADA PPL favors pairs at both 0.6} and \rev{0.7} (Appendix~\ref{app:selective}). \rev{Separately}, \rev{DeepSeek-MoE} at CR \rev{0.5 reaches 63.18} \textit{\rev{Avg7}} \rev{with} $\revmath{G}=\revmath{64}$, \rev{versus 61.90 with} $\revmath{G}=\revmath{2}$ and \rev{63.59 dense} (\rev{Appendix}~\ref{app:moe}).

\subsection{Understanding the Benefits of Sharing}
\label{sec:exp-factorization}\label{sec:exp-budget}
\begin{table}[t]
\caption{\textbf{Factorization choices.} \rev{Average accuracy} (\%, $\uparrow$) at CR $0.4$ with consecutive $G=16$. Pool-wide metrics combine all projections of the same input dimension. Dense coefficients also change dictionary width. Bold marks the highest score.}
\label{tab:factorization-summary}
\ggtabsetup
\tablecolors
\begin{tabularx}{\columnwidth}{@{\hspace{3pt}}lYY@{\hspace{3pt}}}
\toprule
\rowband Variant & Llama-3-8B & Qwen3.5-9B\\
\midrule
\rowours \ggdefault{GeoGroup (default)} & \textbf{66.61} & \textbf{70.58}\\
\midrule
Group-pooled metric & 60.69 & 67.26\\
Pool-wide metric & 57.77 & 66.44\\
\addlinespace[2pt]
Dense coefficients & 38.90 & 33.70\\
Frozen SVD dictionary & 64.74 & 66.67\\
\bottomrule
\end{tabularx}
\end{table}

\paragraph{Activation geometry and sparse representation.}
Holding the grouping fixed, we examine the role of the reconstruction metric and coefficient representation. A group-pooled metric, which pools activation statistics within each group, lowers \textit{\rev{Avg7}} from 66.61/70.58 to 60.69/67.26 on Llama/Qwen. A pool-wide metric, one metric for an entire input-dimension pool, lowers it further to 57.77/66.44 (Table~\ref{tab:factorization-summary}). Retaining projection-specific metrics therefore contributes materially to the quality of the shared representation. Dense coefficients at the same nominal storage budget achieve 38.90/33.70. Under our allocation rule, sparsity allows a wider dictionary, so this comparison captures the combined benefit of sparse codes and the dictionary capacity they enable.\vspace{-3mm}

\paragraph{Capacity enabled by broader sharing.}
Broader sharing increases the capacity available within the budget. In Llama's $d=4096$ pool, the dictionary is already full width at $G=8$, and increasing to $G=16$ raises the mean support fraction $s/d$ from 0.52 to 0.56. In the down-projection pool, both dictionary width and support grow: $k/d$ increases from 0.72 to 0.97 and $s/d$ from 0.29 to 0.39 (Table~\ref{tab:budget}). The selective Qwen3.8-27B experiment exhibits the same mechanism: at CR 0.6, increasing $G$ from 2 to 16 expands both dictionary width and support size by approximately 31\% (Table~\ref{tab:selective}). \vspace{-1mm}

\begin{table}[t]
\caption{\textbf{Capacity enabled by larger groups.} \rev{Average accuracy} (\%, $\uparrow$). ``Budget from'' identifies the group size used to allocate dictionary width and coefficient support. The restricted allocation changes CR, so rows are not all storage matched. Bold marks the highest \rev{average accuracy}.}
\label{tab:capacity-control}
\ggtabsetup
\tablecolors
\begin{tabularx}{\columnwidth}{@{\hspace{3pt}}cYYYY@{\hspace{3pt}}}
\toprule
\rowband $G$ & Budget~from & CR & Llama-3-8B & Qwen3.5-9B\\
\midrule
8 & $G=8$ & 0.40 & 63.93 & 68.83\\
16 & $G=8$ & $\approx0.47$ & 61.46 & 65.87\\
\midrule
16 & $G=16$ & 0.40 & \textbf{66.61} & \textbf{70.58}\\
\bottomrule
\end{tabularx}
\vspace{-3mm}
\end{table}

We probe the role of this additional capacity by restricting $G=16$ groups to the $G=8$ allocation. \textit{\rev{Avg7}} then falls to 61.46/65.87, compared with 63.93/68.83 for standard $G=8$ groups. Restoring the full $G=16$ allocation recovers 5.15/4.71 points (Table~\ref{tab:capacity-control}). The restricted allocation has nominal CR approximately 0.47. This control supports the value of added capacity within larger groups, while membership and budget effects remain coupled.

\subsection{The Role of Group Composition}
\label{sec:exp-grouping}
\begin{table}[t]
\caption{\textbf{Group composition and width-aware alternatives.} CR $0.4$, $G=16$. $\bar\delta$ is the mean within-group distance on Llama, and \rev{average accuracy} is in percent ($\uparrow$). Bold marks the highest \rev{average accuracy}. }
\label{tab:composition}
\ggtabsetup
\tablecolors
\begin{tabularx}{\columnwidth}{@{\hspace{1pt}}lcYY@{\hspace{1pt}}}
\toprule
\rowband Grouping & $\bar\delta$ & Llama-3-8B & Qwen3.5-9B\\
\midrule
HB & 1.44 & 66.27 & 69.88\\
\quad + width balance & 3.72 & 66.76 & 70.53\\
\quad + equalized budgets & 1.44 & 66.43 & \textbf{71.03}\\
\quad + local search & 1.39 & 66.23 & 70.47\\
Balanced spectral clustering & 2.44 & \textbf{67.10} & 70.84\\
\midrule
\ggdefault{Consecutive (default)} & 3.35 & 66.61 & 70.58\\
\bottomrule
\end{tabularx}
\vspace{-2mm}
\end{table}

The capacity analysis suggests that a useful partition should account for output width as well as reconstruction similarity. At $G=16$, Consecutive matches or exceeds the accuracy of HB and Random on both models. Although HB explicitly minimizes an activation-aware distance, its groups in Llama's $d=4096$ pool span widths from $9.3d$ to $35.5d$, compared with $21.5d$ to $27.3d$ for Consecutive (Table~\ref{tab:budget}). The narrower HB groups devote more storage per output column to \rev{the} dictionary and hence receive fewer nonzero coefficients.\vspace{-1mm}

Balancing these widths improves HB's Llama \textit{\rev{Avg7}} from 66.27 to 66.76 even as its mean within-group distance increases from 1.44 to 3.72 (see Table~\ref{tab:composition}). Conversely, local search that reduces the distance yields 66.23. Thus, lower distance alone does not identify better partitions: membership also determines how the representation budget is distributed. Appendix~\ref{app:grouping} examines this interaction and the best-window distance's preference for narrow projections.\vspace{-1mm}

More elaborate selection can yield further gains. Balanced spectral clustering reaches 67.10 on Llama, and equalized-budget HB reaches 71.03 on Qwen. Ordering and remainder handling also affect the outcome (Appendix~\ref{app:consecutive}), while calibration resampling gives standard deviations of up to 0.24 points at $G=16$. Consecutive grouping remains competitive without distance computation or partition search, motivating its use as the default. Width-aware alternatives offer further tuning opportunities.

\subsection{\rev{Preconditioner Ablations}}
\label{sec:exp-solver}
\begin{table}[t]
\caption{\textbf{Dictionary-solver accuracy.} CR $0.4$, consecutive $G=16$, at most 20 PCG iterations per update. Residuals are averaged over dictionary updates, and \rev{average accuracy} is in percent. Bold marks the best value in each column.}
\label{tab:solver-summary}
\ggtabsetup
\setlength{\tabcolsep}{2pt}
\tablecolors
\begin{tabularx}{\columnwidth}{@{\hspace{1pt}}lYYYY@{\hspace{1pt}}}
\toprule
\rowband & \multicolumn{2}{c}{Llama-3-8B} & \multicolumn{2}{c}{Qwen3.5-9B}\\
\cmidrule(lr){2-3}\cmidrule(l){4-5}
\rowband Preconditioner & Residual~$\downarrow$ & \textit{\rev{Avg7}}~$\uparrow$ & Residual~$\downarrow$ & \textit{\rev{Avg7}}~$\uparrow$\\
\midrule
\ggdefault{Jacobi} & $1.1\mathrm{e}{-3}$ & 66.61 & $1.4\mathrm{e}{-3}$ & 70.58\\
Pair-GEVD & $1.1\mathrm{e}{-3}$ & 66.60 & $1.5\mathrm{e}{-3}$ & 70.34\\
Pooled & $\mathbf{1.5e{-4}}$ & \textbf{66.86} & $\mathbf{1.3e{-4}}$ & \textbf{71.24}\\
\bottomrule
\end{tabularx}
\vspace{-2mm}
\end{table}

The \rev{preconditioner ablation} evaluates \rev{three choices within} the \rev{same projection-specific objective}: \rev{Jacobi}, \rev{Pair-GEVD}, \rev{and Pooled}. \rev{We measure dictionary residuals}, \rev{downstream} accuracy, \rev{iteration budgets}, and \rev{compression time to identify which improvements justify their} cost. On Blossom pairs, Pair-GEVD PCG reaches mean relative residuals of $4.1\times10^{-7}$ on Llama and $9.3\times10^{-6}$ on Qwen. Its \textit{\rev{Avg7}} scores, 60.59/60.46, closely match the analytic solver's 60.50/60.57 (Table~\ref{tab:precond}). These measurements complement the conditional algebraic equivalence derived in Sec.~\ref{sec:solver} and Appendix~\ref{app:solver-details}.

For larger groups, preconditioning balances solve accuracy against computational cost. At $G=16$ \rev{and a common 20-iteration cap}, pooled preconditioning reduces residuals \rev{by approximately} $\revmath{7.3}\times$ \rev{on Llama} and $\revmath{10.8}\times$ \rev{on Qwen}, \rev{improving} \textit{\rev{Avg7}} from Jacobi's 66.61/70.58 to 66.86/71.24 (Table~\ref{tab:solver-summary}). \rev{Pair-GEVD instead gives 66.60}/\rev{70.34}: \rev{the pairwise construction does not yield a larger-group advantage at} this \rev{budget}. \rev{Pooled preconditioning} acts only on the solver, \rev{retaining} each projection's metric in the objective. However, \rev{pooled preconditioner} setup exceeds A100 40GB GPU memory for a Llama group of sixteen down projections and requires CPU execution. With the main 20-iteration cap, whole-model Llama compression takes 9.3 hours with pooled preconditioning and 3.9 hours with Jacobi (Table~\ref{tab:solver-cost}). \rev{The iteration-budget ablation reaches the same conclusion}: \rev{pooled preconditioning improves Llama} \textit{\rev{Avg7}} \rev{at caps of 5}, \rev{10}, \rev{20}, \rev{and 50}, \rev{but takes longer than} Jacobi \rev{at the measured caps of 5}, \rev{20}, \rev{and 50 iterations}. \rev{Jacobi is} therefore a \rev{cost-motivated} default, while pooled preconditioning offers additional accuracy when resources permit. Appendix~\ref{app:solver} gives the full solver and iteration-budget studies.

\section{Practical Deployment}
\label{sec:deployment}
We examine storage after accounting for sparse supports, decode performance, and compatibility with quantization. \vspace{-1mm}

\subsection{Effective Storage and Model Size}
To extend the value-count budget in Sec.~\ref{sec:budget} to bits, we account for a bf16 dictionary, packed 14-bit nonzero coefficients, and one support bit per atom and output column. The resulting $16dk+14sM+kM$ bits give a compression ratio relative to dense bf16 weights of
\begin{equation}
\mathrm{CR}_{\mathrm{real}}=\mathrm{CR}+\frac{2s-k}{16d}.
\label{eq:deployment-bits}
\end{equation}
For the consecutive $G=16$ configurations, this accounting gives a targeted-projection CR of 0.402 at nominal CR 0.4. With 16-bit coefficients, the realized CR falls to approximately 0.338, requiring more aggressive factorization to recover the same storage target (Appendix~\ref{app:storage}). The 14-bit accounting requires packed storage.\vspace{-1mm}

Including the components that remain dense gives whole-model storage of 10.45 GB for Llama and 12.34 GB for Qwen, compared with 16.06 and 17.91 GB before compression (Table~\ref{tab:app-system}). The corresponding reductions are approximately 35.0\% and 31.1\%. These estimates describe the packed representation. Serialization overhead and peak runtime memory are separate implementation-dependent quantities.\vspace{-2mm}

\subsection{Compressed-Factor Inference Throughput}
A factorized projection can be evaluated as $\mathbf{x}\mathbf{W}_i\approx(\mathbf{x}\mathbf{D})\mathbf{C}_i$, avoiding reconstruction of the dense matrix $\mathbf{D}\mathbf{C}_i$. Its execution cost depends on sparse-kernel efficiency, support decoding, and activation traffic. In particular, projections receiving different inputs require separate evaluations of $\mathbf{x}\mathbf{D}$, even when they share a dictionary.\vspace{-1mm}

\begin{table}[t]
\caption{\textbf{Full-model decode performance \rev{with MACKO}.} \rev{Only} gate/up \rev{projections are compressed}. Speedup is relative to dense concatenated gate/up (Cat). \rev{Size denotes model-weight} storage \rev{in decimal GB}. Bold marks the best compressed value within each \rev{model}/CR \rev{block}. \rev{Both models use batch size one}. Latency is \rev{reported as} mean $\pm$ \rev{SD} over ten runs. Component \rev{results}: Appendix~\ref{app:decode}.}
\label{tab:decode-summary}
\ggtabsetup
\setlength{\tabcolsep}{2pt}
\tablecolors
\begin{tabular}{@{\hspace{3pt}}lcccc@{\hspace{3pt}}}
\toprule
\rowband & & Latency~$\downarrow$ & Speedup~$\uparrow$ & Size~$\downarrow$\\
\rowband Method & $G$ & (ms/token) & (vs Cat) & (GB)\\
\midrule
\multicolumn{5}{l}{\textbf{\rev{Qwen3.5-9B}}}\\
Dense & -- & $17.471 \pm 0.013$ & -- & 17.91\\
Dense Cat & -- & $17.147 \pm 0.016$ & $1.00\times$ & 17.91\\
\midrule
\multicolumn{5}{@{\hspace{3pt}}l}{\textbf{CR $=0.5$}}\\[2pt]
\rowgeopair GeoPair & 2 & $\mathbf{15.542} \pm 0.001$ & $\mathbf{1.10\times}$ & \textbf{15.26}\\
\rowours Consecutive & 8 & $16.227 \pm 0.001$ & $1.06\times$ & 15.43\\
\rowours \ggdefault{Consecutive} & 16 & $16.353 \pm 0.000$ & $1.05\times$ & 15.46\\
\midrule
\multicolumn{5}{@{\hspace{3pt}}l}{\textbf{CR $=0.6$}}\\[2pt]
\rowgeopair GeoPair & 2 & $\mathbf{15.106} \pm 0.001$ & $\mathbf{1.14\times}$ & \textbf{14.51}\\
\rowours Consecutive & 8 & $15.597 \pm 0.001$ & $1.10\times$ & 14.64\\
\rowours \ggdefault{Consecutive} & 16 & $15.932 \pm 0.001$ & $1.08\times$ & 14.66\\
\midrule
\multicolumn{5}{l}{\textbf{\rev{Qwen3.8-27B}, \rev{CR} $=\revmath{0.6}$}}\\
\rev{Dense} & -- & $\revmath{49.168} \pm \revmath{0.004}$ & -- & \rev{53.79}\\
\rev{Dense Cat} & -- & $\revmath{48.107} \pm \revmath{0.016}$ & $\revmath{1.00}\times$ & \rev{53.79}\\
\rowgeopair \rev{GeoPair} & \rev{2} & $\revmath{\mathbf{40.830}} \pm \revmath{0.001}$ & $\mathbf{\revmath{1.18}\times}$ & \textbf{\rev{41.78}}\\
\rowours \rev{Consecutive} & \rev{16} & $\revmath{43.571} \pm \revmath{0.002}$ & $\revmath{1.10}\times$ & \rev{42.29}\\
\bottomrule
\end{tabular}
\vspace{-3mm}
\end{table}

Selective gate/up compression accelerates full-model decode on both Qwen3.5-9B and Qwen3.8-27B (Table~\ref{tab:decode-summary}), although the gains are smaller than those of the compressed components. For Qwen3.5-9B at batch size one, consecutive $G=16$ achieves 16.353 and 15.932 ms/token at CR 0.5 and 0.6, versus 17.471 dense: $1.07\times$ and $1.10\times$ speedups. Relative to concatenated dense gate/up, the gains are $1.05\times$ and $1.08\times$. The gate/up component alone improves by $1.26\times$ and $1.40\times$.\vspace{-1mm}

On Qwen3.8-27B at batch size one and CR 0.6, consecutive $G=16$ lowers full-model latency from 49.168 to 43.571 ms/token ($1.13\times$ versus dense, $1.10\times$ versus Cat), reaching 23.0 tokens/s. Gate/up latency falls from 18.030 to 12.450 ms/token ($1.45\times$), while full-model weight storage falls by 21\%. These component gains do not translate proportionally to full-model throughput because the remaining computation is unchanged. \vspace{-1mm}

The paired Qwen3.8-27B quality and timing results expose a trade-off. GeoPair reaches 40.830 ms/token and 41.78 GB, compared with 43.571 ms/token and 42.29 GB for consecutive $G=16$. On these same checkpoints, $G=16$ retains 74.04 \textit{Avg7}, 0.90 points above $G=2$ and 0.33 below the dense model's 74.37 (Appendix~\ref{app:selective}). Thus, larger-group compression achieves higher average accuracy, while pairwise compression provides lower latency and weight storage. The configurations differ in realized storage, so this comparison characterizes their measured deployment trade-off rather than isolating group size at equal bytes.\vspace{-1mm}

\subsection{Compatibility with Factor Quantization}
Quantizing the factors can complement the storage savings from sharing. We examine this combination on Qwen3.8-27B after factorizing language-backbone attention and MLP projections at nominal CR 0.4, using $G=8$ for down projections and $G=16$ for the others.
\begin{table}[t]
\caption{\textbf{Compatibility with factor quantization.} Qwen3.8-27B after CR $0.4$ factorization of attention/MLP projections, with \rev{groups of eight} down \rev{projections} and \rev{sixteen projections for the other types}. Factor bits specify dictionary/coefficient precision. Bold marks the best compressed value.}
\label{tab:quantization}
\ggtabsetup
\tablecolors
\begin{tabularx}{\columnwidth}{@{\hspace{3pt}}lYYY@{\hspace{3pt}}}
\toprule
\rowband & Weight~/ & \textit{\rev{Avg7}}~$\uparrow$ & LAMBADA\\
\rowband Representation & factor~bits & (\%) & PPL~$\downarrow$\\
\midrule
Dense & 16 & 74.37 & 3.52\\
\midrule
GeoGroup & 8/6 & 72.67 & \textbf{2.94}\\
GeoGroup & 4/4 & \textbf{73.03} & 3.09\\
\bottomrule
\end{tabularx}
\vspace{-5mm}
\end{table}

With 4-bit dictionary and 4-bit coefficient values, GeoGroup achieves 73.03 \textit{Avg7}, compared with 74.37 for the dense model. The 8-bit/6-bit configuration achieves 72.67 (see Table~\ref{tab:quantization}). Thus, the shared representation remains effective at reduced factor precision in this setting. Total storage also depends on sparse supports, quantization metadata, and uncompressed components. \vspace{-2mm}

\section{Conclusion}
\label{sec:conclusion}
We presented GeoGroup, a training-free framework for sharing sparse dictionaries across groups of projections while preserving their individual activation geometries. Matrix-free PCG makes the resulting multi-term dictionary problem tractable, and a unified preconditioner family connects practical groupwise optimization with exact recovery of the GeoPair dictionary update in the two-member case under matching coefficients and regularization. Our empirical study shows that group composition and representation capacity must be considered together: broader sharing reallocates storage to richer factors, while consecutive grouping achieves competitive accuracy without similarity search. \rev{The accuracy gains extend beyond Llama-3-8B and Qwen3.5-9B to Llama-3.2-1B}, \rev{Phi-4}, \rev{and Gemma-3-4B}, \rev{with additional selective-compression evidence on Qwen3.8-27B and DeepSeek-MoE}. \rev{Preconditioner ablations explain the practical solver choice}: \rev{Jacobi offers low fitting cost}, \rev{while pooled preconditioning can improve accuracy with greater computation and memory}. \rev{GeoGroup's} compatibility with factor quantization and \rev{its} decode \rev{speedups} further support its use \rev{for compressing} pretrained models. \vspace{-2mm}

\section*{Limitations}
Our default configuration uses prescribed grouping and storage-allocation rules. Jointly adapting group composition and coefficient budgets to model structure and layer sensitivity is a promising direction for improving compression quality within a fixed storage budget.\vspace{-1mm}

Deployment benefits depend on hardware, kernels, and storage formats. The throughput \rev{studies cover} batch-size-one decoding with selective gate/up compression. \rev{Evaluating} prefill, \rev{larger batches}, and \rev{runtime memory} would broaden the \rev{deployment} assessment. \vspace{-2mm}

\label{end:main}
\FloatBarrier
\section*{Impact Statement}
The storage and deployment costs of large language models can limit access for researchers, educators, independent developers, and organizations with modest computing resources. GeoGroup advances training-free compression that makes existing pretrained models more compact and supports more efficient inference. Reducing these resource requirements can lower barriers to experimentation and application development, enabling a broader community to work with capable language models. By reusing pretrained models without additional model training and complementing factor quantization, this work contributes practical methods toward more accessible and resource-efficient language-model deployment.

\bibliography{custom}
\bibliographystyle{icml2026}
\clearpage
\appendix
\noindent\textbf{Supplementary material.}
Appendix~\ref{app:setup} specifies the experimental protocol.
Appendix~\ref{app:solver-details} derives and evaluates the dictionary solver.
Appendices~\ref{app:budget-storage} and~\ref{app:grouping-all} examine capacity and group composition.
Appendices~\ref{app:further} and~\ref{app:mm} provide extended language and multimodal results,
and Appendix~\ref{app:system} details storage, quantization, and deployment.

\noindent \rev{In supplementary tables}, \rev{bold marks the best result within each comparison specified in the caption}, \rev{including ties}. \rev{Dense references are excluded unless stated otherwise}.

\FloatBarrier
\section{Experimental Protocol}
\label{app:setup}
This appendix specifies the common setup for the quality experiments in Section~\ref{sec:experiments}. We distinguish calibration variability from grouping variability and explain the task-specific metrics used to interpret the results. Selective compression, multimodal calibration, and decode measurements use the settings stated in their respective appendices.

\subsection{Models, Calibration, and Baselines}
\label{app:projection-pools}
Table~\ref{tab:app-config} summarizes the default configuration. Text-only factorization experiments use 256 FineWeb sequences of 1,024 tokens. Methods evaluated under this protocol compress the same projection set and use the same calibration data. Their optimization budgets and representations differ as specified in the table. ROCKET and COMPOT are rerun under the same conditions as GeoGroup and GeoPair. Table~\ref{tab:main} reports per-task rerun scores at CR 0.4 and aggregate accuracy and LAMBADA perplexity at CR 0.5.

Groups are constructed within pools of equal input dimension. In Llama-3-8B, query, key, value, output, gate, and up projections form a pool with $d=4096$, while down projections form a second pool with $d=14336$. Consecutive groups of sixteen in the first pool span parts of three or four adjacent layers. Qwen3.5-9B similarly groups compatible softmax-attention, linear-attention, gate, and up projections separately from down projections. Embeddings, output heads, and vision encoders remain dense.

The single-projection dense-code configuration closely reproduces SVD-LLM accuracy: with per-projection whitening and one SVD, it achieves 33.85/41.69 \textit{Avg7} on Llama/Qwen, compared with 33.82/41.67 for SVD-LLM.

\begin{table}[!tb]
\caption{Default configuration of \method{} and the baselines.}
\label{tab:app-config}
\centering
\footnotesize
\setlength{\tabcolsep}{2pt}
\tablecolors
\begin{tabularx}{\columnwidth}{@{\hspace{3pt}}>{\raggedright\arraybackslash}p{2.3cm}>{\raggedright\arraybackslash}X@{\hspace{3pt}}}
\toprule
\rowband Item & Setting \\
\midrule
Models & Llama-3-8B, Llama-3.2-1B, Qwen3.5-9B, Phi-4, Gemma-3-4B, Qwen3.8-27B \\
Calibration & 256 FineWeb sequences $\times$ 1,024 tokens, seed 2023 \\
Compressed weights & backbone attention and MLP projections, unless a selective scope is specified \\
Pools & equal input dimension $d$ \\
Budget & $d\natoms+\nnz M=(1-\mathrm{CR})\,dM$, $\kappa=\natoms/\nnz=2.5$, $\natoms\le\min(d,M)$ \\
Initialization & top-$\natoms$ left singular vectors of $[\mathbf{W}_1,\dots,\mathbf{W}_G]$ \\
$\mathbf{D}$-step & PCG, Jacobi, $\le20$ iterations, rel.\ residual $10^{-4}$, warm start \\
$\mathbf{C}$-step & two consecutive hard thresholding pursuit (HTP) steps, each with a 10-CG support refit \\
Rounds & 30, stop after 10 without improvement \\
Kept iterate & lowest value of Eq.~\eqref{eq:objective} \\
Precision & float32 factors, bf16 reconstructed weights \\
HB at $G=2$ & Blossom matching \\
Random grouping & 3--7 seeds per setting \\
Evaluation & LM Evaluation Harness 0.4.9.2, seven zero-shot tasks, metric conventions in Appendix~\ref{app:evaluation-details} \\
Factorization hardware & NVIDIA A100 40GB, two GPUs for model compression, one for solver diagnostics \\
\midrule
\geopair{} & Blossom pairs, analytic FP64 update, 50 rounds \\
Basis Sharing & official code, $G=2$ ($G=16$ in Table~\ref{tab:app-mm}) \\
SVD-LLM & whitening-only variant \\
\bottomrule
\end{tabularx}

\end{table}

\subsection{Evaluation Metrics}
\label{app:evaluation-details}
We report normalized accuracy for tasks that support this metric and standard accuracy for the remaining tasks. \textit{Avg7} is the unweighted mean of the seven task scores, computed before rounding. LAMBADA, WikiText-2, and C4 perplexities provide complementary measures of language modeling quality.

\paragraph{Length normalization.}
For ARC, normalized accuracy selects answers using log-likelihood divided by answer length in characters. Table~\ref{tab:app-arc} gives both metrics. Under standard accuracy, consecutive $G=16$ remains below the dense reference on ARC-Challenge for all three models, while exceeding it on ARC-Easy for Qwen and Phi-4. Thus, some improvements over dense models in the main tables depend on the scoring convention and should be interpreted using the corresponding metric.
\begin{table}[!tb]
\caption{ARC accuracy (\%) without and with length normalization for the dense models and Consecutive $G=16$ at CR $0.4$. \rev{Bold marks the higher accuracy within each model and scoring convention}, \rev{including the dense reference}.}
\label{tab:app-arc}
\centering
\scriptsize
\setlength{\tabcolsep}{2pt}
\tablecolors
\begin{tabularx}{\columnwidth}{@{\hspace{3pt}}lYYYY@{\hspace{3pt}}}
\toprule
\rowband & \multicolumn{2}{c}{ARC-Easy} & \multicolumn{2}{c}{ARC-Challenge} \\
\cmidrule(lr){2-3}\cmidrule(lr){4-5}
\rowband Variant & Accuracy & \makecell{Normalized\\accuracy} & Accuracy & \makecell{Normalized\\accuracy} \\
\midrule
\multicolumn{5}{@{\hspace{3pt}}l}{\textit{Llama-3-8B}} \\
\quad \dense{Dense} & \dense{\textbf{80.7}} & \dense{\textbf{77.6}} & \dense{\textbf{51.7}} & \dense{\textbf{54.0}} \\
\quad Consecutive & 76.9 & 76.1 & 45.9 & 49.3 \\
\multicolumn{5}{@{\hspace{3pt}}l}{\textit{Qwen3.5-9B}} \\
\quad \dense{Dense} & \dense{81.3} & \dense{74.3} & \dense{\textbf{54.4}} & \dense{55.9} \\
\quad Consecutive & \textbf{81.6} & \textbf{81.0} & 52.7 & \textbf{57.6} \\
\multicolumn{5}{@{\hspace{3pt}}l}{\textit{Phi-4}} \\
\quad \dense{Dense} & \dense{81.6} & \dense{74.0} & \dense{\textbf{55.5}} & \dense{56.0} \\
\quad Consecutive & \textbf{83.3} & \textbf{81.5} & 54.9 & \textbf{58.5} \\
\bottomrule
\end{tabularx}

\end{table}

\paragraph{Interpreting perplexity.}
On Llama-3-8B, consecutive $G=16$ improves LAMBADA perplexity from 3.13 to 2.78 while increasing WikiText-2 perplexity from 6.14 to 9.07. These datasets probe different prediction settings, so an improvement on LAMBADA does not establish an improvement in language modeling across datasets. Appendix~\ref{app:bench} complements these measurements with reasoning and generation tasks.

\subsection{Calibration Sensitivity}
\label{app:seeds}
To separate calibration variability from partition variability, Table~\ref{tab:app-seeds} holds each partition fixed and resamples the calibration sequences. The standard deviation of \textit{Avg7} is at most 0.33 points overall and 0.24 at $G=16$. This variation provides context for the smaller differences among grouping and solver variants. It is substantially below the gains over GeoPair in the main comparison. On Llama-3-8B, reducing the calibration set from 256 to 128 or 64 sequences gives 66.45 or 65.85 \textit{Avg7}, respectively, compared with 66.61 for the default.
\begin{table}[!tb]
\caption{\rev{Sensitivity to calibration sampling at} CR $0.4$ unless noted. \rev{Average accuracy is reported} as mean $\pm$ SD, with the number of seeds in \rev{parentheses}. The grouping is fixed, and only the 256 calibration sequences are redrawn. Blossom pairs use Jacobi PCG and 30 rounds, and \rev{individual runs} score 59.78, 60.18, and 60.29 on Llama-3-8B. The SD is at most 0.33 points. \rev{Bold marks the highest mean at CR 0.4}. \rev{The CR 0.2 and 0.5 rows are separate settings}.}
\label{tab:app-seeds}
\centering
\footnotesize
\setlength{\tabcolsep}{2pt}
\tablecolors
\begin{tabularx}{\columnwidth}{@{\hspace{3pt}}lYY@{\hspace{3pt}}}
\toprule
\rowband Run & Llama-3-8B & Qwen3.5-9B \\
\midrule
Blossom pairs, Jacobi & \sd{60.08}{0.27}~(3) & \sd{57.77}{0.06}~(2) \\
Consecutive $G{=}2$ & \sd{61.32}{0.28}~(3) & \sd{61.48}{0.02}~(2) \\
Consecutive $G{=}4$ & \sd{62.89}{0.20}~(3) & \sd{64.81}{0.04}~(3) \\
Consecutive $G{=}8$ & \sd{64.04}{0.09}~(3) & \sd{68.82}{0.01}~(2) \\
Consecutive $G{=}16$ & \sd{66.62}{0.02}~(3) & \sd{\textbf{70.48}}{0.14}~(2) \\
Consecutive $G{=}16$, CR 0.2 & \sd{69.31}{0.03}~(3) & \sd{72.27}{0.08}~(2) \\
Consecutive $G{=}16$, CR 0.5 & \sd{61.16}{0.19}~(3) & \sd{64.45}{0.10}~(2) \\
Input family $G{=}4$ & \sd{63.16}{0.32}~(3) & \sd{64.91}{0.26}~(3) \\
Input family $G{=}16$ & \sd{\textbf{66.93}}{0.08}~(3) & \sd{70.32}{0.24}~(3) \\
Projection $G{=}8$ & \sd{63.12}{0.26}~(3) & \sd{69.27}{0.33}~(2) \\
\bottomrule
\end{tabularx}

\end{table}

\section{Dictionary Optimization and Preconditioning}
\label{app:solver-details}

We derive the matrix-free dictionary update and its preconditioner family, establish the pairwise connection to GeoPair, and evaluate the accuracy--cost trade-off of the solver.

\subsection{Optimization Derivations and Preconditioners}
\label{app:optimization-derivations}

This appendix provides the regularized activation metric, sparse coefficient updates, properties of the dictionary normal equations, and preconditioners for the matrix-free solver in Section~\ref{sec:solver}.

\paragraph{Activation metric and regularization.}
For a projection with calibration inputs $\mathbf{X}$ and approximation $\widehat{\mathbf{W}}$, the unregularized Gram factor $\mathbf{L}^\top\mathbf{L}=\mathbf{X}^\top\mathbf{X}$ satisfies
\begin{equation}
\|\mathbf{X}(\mathbf{W}-\widehat{\mathbf{W}})\|_F
=\|\mathbf{L}(\mathbf{W}-\widehat{\mathbf{W}})\|_F.
\label{eq:whitening}
\end{equation}
When the Gram matrix is singular, we instead factor
$\mathbf{L}_i^\top\mathbf{L}_i=\mathbf{X}_i^\top\mathbf{X}_i+\eta_i\mathbf{I}$,
where $\eta_i>0$ is small. Writing $\mathbf{E}_i=\mathbf{W}_i-\mathbf{D}\mathbf{C}_i$, the regularized objective then obeys
\begin{equation}
\|\mathbf{L}_i\mathbf{E}_i\|_F^2
=\|\mathbf{X}_i\mathbf{E}_i\|_F^2+\eta_i\|\mathbf{E}_i\|_F^2.
\label{eq:ridge-error}
\end{equation}
The same $\mathbf{L}_i$ is used in the numerator and normalization of Eq.~\eqref{eq:objective}, so the dictionary normal equation corresponds to the stated activation-regularized loss.

\paragraph{Sparse coefficient update.}
With the dictionary fixed, each column $\mathbf{w}$ of $\mathbf{W}_i$ has the constrained least-squares problem
\begin{equation}
\min_{\|\mathbf{c}\|_0\le\nnz}
\|\mathbf{L}_i(\mathbf{w}-\mathbf{D}\mathbf{c})\|_2^2.
\label{eq:coefficient-update}
\end{equation}
Define $\mathbf{H}_i=\mathbf{D}^\top\mathbf{L}_i^\top\mathbf{L}_i\mathbf{D}+\epsilon\mathbf{I}$ and $\mathbf{r}=\mathbf{D}^\top\mathbf{L}_i^\top\mathbf{L}_i\mathbf{w}$, where $\epsilon=10^{-6}$ in our implementation. The $\epsilon$ term stabilizes the coefficient subproblem and is an additional coefficient penalty, not part of the activation ridge in Eq.~\eqref{eq:ridge-error}. Initial coefficients solve the dense normal equations $\mathbf{H}_i\mathbf{c}=\mathbf{r}$. Each alternating round applies \rev{two consecutive} hard thresholding pursuit (HTP) \rev{steps}~\citep{foucart2011htp} \rev{per coefficient update}. \rev{In each step}, form $\widetilde{\mathbf{c}}=\mathbf{c}+(\mathbf{r}-\mathbf{H}_i\mathbf{c})$, select the $\nnz$ largest-magnitude entries as support $\Omega$, and refit by solving $(\mathbf{H}_i)_{\Omega\Omega}\mathbf{c}_{\Omega}=\mathbf{r}_{\Omega}$ with $\mathbf{c}_{\Omega^c}=\mathbf{0}$. \rev{Each support refit uses} 10 iterations of batched conjugate gradients. \rev{The second HTP step starts from the coefficients refitted in the first}. This inexact update uses a unit proposal step, and its behavior is assessed empirically rather than through an HTP convergence guarantee.

\paragraph{Dictionary normal equations.}
For fixed coefficients, define
\begin{equation}
\begin{aligned}
\alpha_i &= \|\mathbf{L}_i\mathbf{W}_i\|_F^{-2}, &
\mathbf{A}_i &= \alpha_i\mathbf{L}_i^\top\mathbf{L}_i,\\
\mathbf{B}_i &= \mathbf{C}_i\mathbf{C}_i^\top, &
\mathbf{K} &= \sum_{i\in\mathcal S}\mathbf{A}_i\mathbf{W}_i\mathbf{C}_i^\top.
\end{aligned}
\label{eq:operator-factors}
\end{equation}
Differentiating the quadratic objective gives
$\nabla_{\mathbf{D}}\mathcal L=2\big(\sum_i\mathbf{A}_i\mathbf{D}\mathbf{B}_i-\mathbf{K}\big)$, yielding Eq.~\eqref{eq:sylvester}. The operator is self-adjoint and positive semidefinite in the Frobenius inner product:
\begin{equation}
\langle\mathbf{Z},\mathcal A(\mathbf{Z})\rangle_F
=\sum_{i\in\mathcal S}\alpha_i\|\mathbf{L}_i\mathbf{Z}\mathbf{C}_i\|_F^2\ge0.
\label{eq:operator-psd}
\end{equation}
When all $\mathbf{L}_i$ are nonsingular, $\mathcal A$ is positive definite if and only if the concatenated coefficient matrix $[\mathbf{C}_i]_{i\in\mathcal S}$ has full row rank. Otherwise the objective remains convex in $\mathbf{D}$ but its minimizer need not be unique. Standard PCG applies in the full-rank case. We floor preconditioner diagonal entries at $10^{-8}$ for numerical stability. This protects the preconditioner but does not alter the rank of the dictionary operator.

Vectorization yields the system
$\big(\sum_i\mathbf{B}_i\otimes\mathbf{A}_i\big)\operatorname{vec}(\mathbf{D})=\operatorname{vec}(\mathbf{K})$.
We never assemble this $d\natoms$-dimensional coefficient matrix: PCG evaluates Eq.~\eqref{eq:matrix-free-operator}, warm-starting from the preceding dictionary iterate. Our main experiments use at most 20 PCG iterations per dictionary update and stop earlier when $\|\mathbf{K}-\mathcal A(\mathbf{D})\|_F\le10^{-4}\|\mathbf{K}\|_F$. We alternate for at most 30 rounds, stop after 10 rounds without objective improvement, and keep the best iterate.

\paragraph{Rotated diagonal preconditioners.}
For invertible bases $\mathbf{U}\in\mathbb{R}^{d\times d}$ and $\mathbf{V}\in\mathbb{R}^{\natoms\times\natoms}$, write $\widehat{\mathbf{A}}_i=\mathbf{U}^\top\mathbf{A}_i\mathbf{U}$ and $\widehat{\mathbf{B}}_i=\mathbf{V}^\top\mathbf{B}_i\mathbf{V}$. Our family of preconditioners is
\begin{equation}
\begin{aligned}
\mathcal M_{\mathbf{U}\mathbf{V}}^{-1}(\mathbf{R})
&=\mathbf{U}\left[(\mathbf{U}^\top\mathbf{R}\mathbf{V})\oslash\boldsymbol{\Delta}\right]\mathbf{V}^\top,\\
\Delta_{ab}&=\sum_{i\in\mathcal S}(\widehat{\mathbf{A}}_i)_{aa}(\widehat{\mathbf{B}}_i)_{bb},
\end{aligned}
\label{eq:rotated-jacobi}
\end{equation}
where $\oslash$ denotes elementwise division and $\boldsymbol{\Delta}$ is the diagonal of the transformed Kronecker system, reshaped as a matrix. Jacobi uses $\mathbf{U}=\mathbf{I}$, $\mathbf{V}=\mathbf{I}$. The pooled variant uses eigenvectors of $\sum_i\mathbf{A}_i$ and $\sum_i\mathbf{B}_i$. These rotations affect only the preconditioner. The objective retains each projection's own metric.

\paragraph{Pair-GEVD and the pairwise closed form.}
For a reference pair $(p,q)$, suppose generalized eigendecomposition provides bases $\mathbf{P},\mathbf{Q}$ satisfying
\begin{equation}
\begin{aligned}
\mathbf{P}^\top\mathbf{A}_p\mathbf{P}&=\mathbf{I}, &
\mathbf{P}^\top\mathbf{A}_q\mathbf{P}&=\boldsymbol{\Lambda},\\
\mathbf{Q}^\top\mathbf{B}_p\mathbf{Q}&=\mathbf{I}, &
\mathbf{Q}^\top\mathbf{B}_q\mathbf{Q}&=\boldsymbol{\Theta}.
\end{aligned}
\label{eq:pair-gevd}
\end{equation}
For $G=2$ with the required positive-definite factors, setting $\mathbf{D}=\mathbf{P}\mathbf{Y}\mathbf{Q}^\top$ reduces the Sylvester equation to $\mathbf{Y}+\boldsymbol{\Lambda}\mathbf{Y}\boldsymbol{\Theta}=\mathbf{P}^\top\mathbf{K}\mathbf{Q}$. Writing $\boldsymbol{\lambda}=\operatorname{diag}(\boldsymbol{\Lambda})$ and $\boldsymbol{\theta}=\operatorname{diag}(\boldsymbol{\Theta})$, the solution is
\begin{equation}
\mathbf{D}=\mathbf{P}\left[
(\mathbf{P}^\top\mathbf{K}\mathbf{Q})\oslash
(\mathbf{1}_{d}\mathbf{1}_{\natoms}^\top+\boldsymbol{\lambda}\boldsymbol{\theta}^\top)
\right]\mathbf{Q}^\top.
\label{eq:pair-closed-form}
\end{equation}
This is the pairwise update of \geopair{} when the coefficients and dictionary regularization match. In its regularized formulation, replace every $\mathbf B_i$ in the operator and its generalized eigendecomposition by $\widetilde{\mathbf B}_i=\mathbf C_i\mathbf C_i^\top+\epsilon_D\mathbf I$. This is equivalent to adding $\epsilon_D\sum_i\alpha_i\|\mathbf L_i\mathbf D\|_F^2$ to the objective. Regularizing only the preconditioner instead leaves the original operator unchanged and does not give exact recovery of this regularized solution. In exact arithmetic, the corresponding preconditioner solves the pair system in at most one PCG iteration. For $G>2$, additional terms prevent this exact diagonalization in general, but the reference-pair bases still define a preconditioner. We test up to four reference pairs, reject candidates whose wider member has $m_p<\natoms$ (a necessary dimensional screen, not a sufficient rank test), and select among the rest using a bound on the transformed system's off-diagonal mass. Otherwise, we fall back to Jacobi.

\paragraph{A bound on preconditioner quality.}
Let $\widehat{\mathbf{H}}=\sum_i\widehat{\mathbf{B}}_i\otimes\widehat{\mathbf{A}}_i$, let $\mathbf{D}_{\Delta}$ contain its diagonal, and define $\rho=\|\widehat{\mathbf{H}}-\mathbf{D}_{\Delta}\|_F/\min_{ab}\Delta_{ab}$.
\begin{proposition}\label{prop:precond}
If all $\Delta_{ab}>0$ and $\rho<1$, the eigenvalues of the preconditioned operator lie in $[1-\rho,1+\rho]$. If all $\widehat{\mathbf{A}}_i$ and $\widehat{\mathbf{B}}_i$ are diagonal, the preconditioner is exact and PCG terminates after at most one iteration in exact arithmetic.
\end{proposition}
\noindent\textit{Proof.}
The transformed Kronecker operator is $\widehat{\mathbf{H}}$. Its diagonally preconditioned form is similar to $\mathbf{I}+\mathbf{D}_{\Delta}^{-1/2}(\widehat{\mathbf{H}}-\mathbf{D}_{\Delta})\mathbf{D}_{\Delta}^{-1/2}$. The perturbation has spectral norm at most $\rho$, giving the eigenvalue interval. If every transformed factor is diagonal, $\widehat{\mathbf{H}}=\mathbf{D}_{\Delta}$, so the preconditioned operator is the identity.\hfill$\square$

A computable upper bound follows from the decomposition of each Kronecker term into its diagonal and off-diagonal parts:
\begin{equation}
\begin{aligned}
\rho\min_{ab}\Delta_{ab}\le\sum_{i\in\mathcal S}\Big(&
\|\operatorname{off}(\widehat{\mathbf{B}}_i)\|_F^2
\|\widehat{\mathbf{A}}_i\|_F^2\\
&+\|\operatorname{diag}(\widehat{\mathbf{B}}_i)\|_F^2
\|\operatorname{off}(\widehat{\mathbf{A}}_i)\|_F^2
\Big)^{1/2}.
\end{aligned}
\label{eq:precond-bound}
\end{equation}
Here $\operatorname{off}$ zeros a matrix diagonal and $\operatorname{diag}$ zeros its off-diagonal entries. For Pair-GEVD with exact reference-factor diagonalization, only non-reference members contribute. Candidate reference pairs include the widest projection with its widest eligible partner, as well as first/last, first/middle, and middle/last members in layer order. The bound ranks candidates. Because it is sufficient rather than necessary, a value above one does not imply solver failure.

Generalized eigendecompositions use FP64 and PCG uses FP32. A ridge applied only during preconditioner construction can perturb the exact pairwise inverse. Exact GeoPair recovery requires the matched operator and regularization described above. One matrix-free operator application costs $O(Gd^2\natoms+d\natoms M)$. Jacobi requires $O(d\natoms)$ work per preconditioner application, whereas a rotated preconditioner requires $O(d^2\natoms+d\natoms^2)$.

\subsection{Solver Accuracy and GeoPair Recovery}
\label{app:solver}
\paragraph{Recovery on pairs.}
Table~\ref{tab:precond}(a) compares the analytic GeoPair update with PCG on the same Blossom pairs and with the same 50-round budget. Pair-GEVD requires one PCG step on average on Llama and reaches mean relative residuals of $4.1\times10^{-7}$ on Llama and $9.3\times10^{-6}$ on Qwen. Its downstream scores differ from the analytic solver by 0.09 and 0.11 points, respectively. These measurements support the algebraic connection established above while distinguishing the exact dictionary subproblem from the finite-precision alternating fit.

\paragraph{Preconditioning beyond pairs.}
Table~\ref{tab:precond}(b) evaluates consecutive groups with 30 alternating rounds. Pooled rotations generally reduce residuals more strongly than Jacobi, while Llama \textit{\rev{Avg7}} differs by at most 0.3 points among the preconditioners for $G\ge4$. Pair-GEVD remains a preconditioner for larger groups but no longer diagonalizes the full operator. The consecutive-pair results also illustrate that solver residuals and downstream accuracy measure different properties: a smaller residual need not yield a higher task score. The Blossom and consecutive panels use different partitions and round budgets and should be interpreted within their respective settings.
\begin{table*}[!t]
\caption{\textbf{Solver accuracy and recovery of the pairwise update.} CR $0.4$. Blossom pairs use 50 alternating rounds, and consecutive groups use 30. PCG is capped at 20 iterations per dictionary update. \rev{Entries marked} $^\dagger$ \rev{use a 10-iteration cap}. Steps and residuals are means over dictionary updates. Calibration error is scaled by $10^2$, and \rev{average accuracy} is in percent. Dashes denote unavailable measurements or inapplicable iteration counts. Pair-GEVD falls back to Jacobi on down projections for $G\ge4$. \rev{Bold marks the lowest residual}/\rev{calibration error and highest average accuracy within each grouping}/$\revmath{G}$ \rev{block}, \rev{including the closed-form reference in panel} (\rev{a}).}
\label{tab:precond}
\centering\footnotesize
\renewcommand{\arraystretch}{1.08}
\setlength{\tabcolsep}{3pt}
\tablecolors
\begin{tabularx}{\textwidth}{@{\hspace{3pt}}clYYYYYYY@{\hspace{3pt}}}
\toprule
\rowband & & \multicolumn{4}{c}{Llama-3-8B} & \multicolumn{3}{c}{Qwen3.5-9B}\\
\cmidrule(lr){3-6}\cmidrule(l){7-9}
\rowband $G$ & Dictionary solver & PCG~steps & Residual~$\downarrow$ & Calib.~error~$\downarrow$ & \textit{\rev{Avg7}}~$\uparrow$ & Residual~$\downarrow$ & Calib.~error~$\downarrow$ & \textit{\rev{Avg7}}~$\uparrow$\\
\midrule
\multicolumn{9}{@{\hspace{3pt}}l}{\textbf{(a) Blossom pairs: comparison with GeoPair, 50 rounds}}\\[2pt]
\rowgeopair 2 & Closed form (GeoPair) & -- & -- & \textbf{2.565} & 60.50 & -- & \textbf{2.208} & \textbf{60.57}\\
2 & Jacobi & 19.6 & 2.0e-3 & 2.594 & 60.08 & 1.4e-3 & 2.266 & 56.97\\
2 & Pair-GEVD & 1.0 & \textbf{4.1e-7} & \textbf{2.565} & \textbf{60.59} & \textbf{9.3e-6} & 2.217 & 60.46\\
2 & Pooled & 7.1 & 1.1e-4 & 2.566 & 60.51 & 1.3e-4 & 2.220 & 59.86\\
\midrule
\multicolumn{9}{@{\hspace{3pt}}l}{\textbf{(b) Consecutive groups: scaling beyond pairs, 30 rounds}}\\[2pt]
2 & Jacobi & 20.0 & 2.3e-3 & 2.369 & 61.36 & 1.6e-3 & 2.279 & 61.49\\
2 & Pair-GEVD & 2.4 & \textbf{3.1e-5} & \textbf{2.316} & 60.66 & -- & -- & --\\
2 & Pooled & 7.1 & 9.0e-5 & 2.318 & \textbf{61.63} & -- & -- & --\\
\addlinespace[3pt]
4 & Jacobi & 20.0 & 2.1e-3 & 1.749 & 62.66 & 1.9e-3 & 1.530 & 64.81\\
4 & Pair-GEVD & 13.5 & 1.1e-3 & 1.738 & 62.60 & 1.1e-3 & 1.514 & 64.99\\
4 & Pooled & 13.3 & \textbf{1.4e-4} & \textbf{1.723} & \textbf{62.78} & \textbf{1.3e-4} & \textbf{1.490} & \textbf{65.06}\\
\addlinespace[3pt]
8 & Jacobi & 20.0 & 1.5e-3 & 1.340 & \textbf{63.93} & 1.9e-3 & \textbf{1.154} & \textbf{68.83}\\
8 & Pair-GEVD & 20.0 & 1.7e-3 & 1.381 & 63.64 & \textbf{1.8e-3} & 1.200 & 68.45\\
8 & Pooled$^\dagger$ & 9.5 & \textbf{2.9e-4} & \textbf{1.333} & 63.88 & -- & -- & --\\
\addlinespace[3pt]
16 & Jacobi & 20.0 & 1.1e-3 & 1.073 & 66.61 & 1.4e-3 & 0.940 & 70.58\\
16 & Pair-GEVD & 20.0 & 1.1e-3 & 1.089 & 66.60 & 1.5e-3 & 0.971 & 70.34\\
16 & Pooled & 13.2 & \textbf{1.5e-4} & \textbf{1.070} & \textbf{66.86} & \textbf{1.3e-4} & \textbf{0.936} & \textbf{71.24}\\
\bottomrule
\end{tabularx}
\end{table*}

\subsection{Computational Cost and Solver Budget}
\paragraph{Whole-model accuracy--cost trade-off.}
Table~\ref{tab:solver-cost}(a) varies the PCG cap at $G=16$. Increasing it from 5 to 50 improves Llama \textit{\rev{Avg7}} by 0.56 points with Jacobi and 0.16 with pooled preconditioning. The default 20-iteration Jacobi configuration balances this gain against fitting time. At that cap, pooled preconditioning improves \textit{\rev{Avg7}} from 66.61 to 66.86 but increases whole-model compression time from 3.9 to 9.3 hours.

\paragraph{Wide-projection memory and runtime.}
Table~\ref{tab:solver-cost}(b) isolates groups of down projections, using one GPU and allowing up to 50 PCG iterations per update. Pooled preconditioning is faster than Jacobi through $G=8$. At $G=16$, its setup exceeds the A100 40GB GPU memory budget and runs on the CPU, increasing the time for 30 rounds from 2.89 to 4.80 hours. Pair-GEVD falls back to Jacobi for these groups from $G=4$ because no member is as wide as the dictionary. These per-group measurements explain the practical advantage of Jacobi for wide groups. Their times are distinct from the whole-model measurements in panel (a).
\begin{table*}[!t]
\caption{\textbf{Solver budget and computational cost.} Llama-3-8B, CR $0.4$, 30 alternating rounds. Panel (a) measures whole-model compression at different PCG caps, and panel (b) measures individual groups of down projections with a cap of 50 and relative-residual tolerance $10^{-4}$. All times are in hours. \rev{Average accuracy} is in percent, and peak GPU memory is in GB. The panels use different workloads and GPU counts. \rev{Bold marks the best average accuracy and lowest available time at each cap in} (\rev{a}), \rev{and the lowest time at each} $\revmath{G}$ \rev{in} (\rev{b}).}
\label{tab:solver-cost}
\centering\footnotesize
\renewcommand{\arraystretch}{1.08}
\begin{minipage}[t]{0.485\textwidth}
\centering
\textbf{(a) Whole-model fitting: consecutive $G=16$}\\[2pt]
Two A100 40GB GPUs\\[5pt]
\setlength{\tabcolsep}{2pt}
\tablecolors
\begin{tabularx}{\linewidth}{@{\hspace{3pt}}cYYYYYY@{\hspace{3pt}}}
\toprule
\rowband & \multicolumn{2}{c}{Jacobi} & \multicolumn{2}{c}{Pair-GEVD} & \multicolumn{2}{c}{Pooled}\\
\cmidrule(lr){2-3}\cmidrule(lr){4-5}\cmidrule(l){6-7}
\rowband PCG cap & \textit{\rev{Avg7}}$\uparrow$ & Time & \textit{\rev{Avg7}}$\uparrow$ & Time & \textit{\rev{Avg7}}$\uparrow$ & Time\\
\midrule
5 & 66.27 & \textbf{1.9} & 66.44 & 2.0 & \textbf{66.72} & 8.3\\
10 & 66.58 & \textbf{2.6} & 66.55 & \textbf{2.6} & \textbf{66.64} & --\\
20 & 66.61 & \textbf{3.9} & 66.60 & 4.0 & \textbf{66.86} & 9.3\\
50 & 66.83 & \textbf{7.8} & 66.79 & 8.1 & \textbf{66.88} & 10.3\\
\bottomrule
\end{tabularx}
\par\vspace{4pt}\raggedright
\rev{A dash} denotes an unavailable timing.
\end{minipage}\hfill
\begin{minipage}[t]{0.485\textwidth}
\centering
\textbf{(b) One group of $G$ down projections}\\[2pt]
One A100 40GB GPU, $d=14336$, FP32\\[5pt]
\setlength{\tabcolsep}{2pt}
\tablecolors
\begin{tabularx}{\linewidth}{@{\hspace{3pt}}cYcYYYY@{\hspace{3pt}}}
\toprule
\rowband & Jacobi & Pair-GEVD & \multicolumn{4}{c}{Pooled}\\
\cmidrule(l){4-7}
\rowband $G$ & Time & Time & Time & Steps & Peak & Setup\\
\midrule
2 & 0.22 & \textbf{0.04} & 0.10 & 11.7 & 18.0 & GPU\\
4 & 0.53 & 0.53 & \textbf{0.38} & 22.1 & 21.3 & GPU\\
8 & 1.25 & 1.25 & \textbf{0.82} & 21.1 & 27.7 & GPU\\
16 & \textbf{2.89} & 2.92 & 4.80 & 25.1 & 40.7 & CPU\\
\bottomrule
\end{tabularx}
\par\vspace{4pt}\raggedright
Groups start at layer 16. Steps are means per update. Setup device and peak GPU memory refer to pooled preconditioning. Pair-GEVD falls back to Jacobi for $G\ge4$.
\end{minipage}
\end{table*}

\section{Representation Capacity and Storage Allocation}
\label{app:budget-storage}
This appendix develops the storage mechanism underlying Section~\ref{sec:exp-budget}. We first derive how sharing reallocates the budget, then examine the capacity control and the factorization choices that determine how this capacity is used.

\begin{figure}[!tb]
\centering
\input{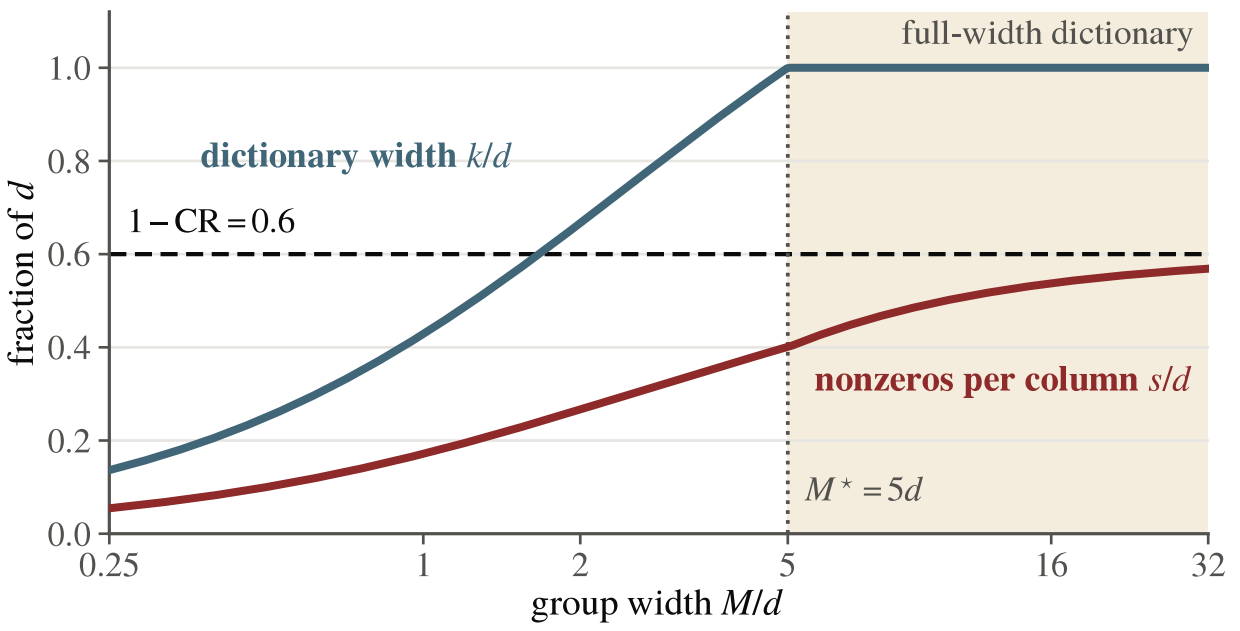}
\caption{Allocation at CR $0.4$ and $\kappa=2.5$. Sharing first permits a wider dictionary. After saturation, additional group width increases the support per coefficient column toward $(1-\mathrm{CR})d$.}
\label{fig:budget}
\end{figure}

\subsection{Group-Level Allocation and Dictionary Saturation}
\label{app:allocation-details}

\paragraph{Allocation under a fixed budget.}
For a group with input dimension $d$, total output width $M$, dictionary width $\natoms$, and $\nnz$ nonzeros per coefficient column, the nominal retained-value budget is $d\natoms+\nnz M=(1-\mathrm{CR})dM$ (Eq.~\eqref{eq:budget}). Before applying the cap $\natoms\le\min(d,M)$, the allocation ratio $\kappa=\natoms/\nnz$ gives
\begin{equation}
\natoms_0=\frac{(1-\mathrm{CR})dM}{d+M/\kappa}.
\label{eq:allocation-uncapped}
\end{equation}
We use $\kappa=2.5$ as the initial ratio, set $\natoms=\min(\natoms_0,d,M)$, and allocate the remaining budget to $\nnz$ as in Eq.~\eqref{eq:support-budget}. Integer rounding is applied after this allocation. If $\natoms=d$, the realized ratio $\natoms/\nnz$ may be smaller than $\kappa$.

\paragraph{Full-width saturation.}
When $1-\mathrm{CR}>1/\kappa$, the dictionary reaches its maximum width at
\begin{equation}
M\ge M^\star=\frac{d}{1-\mathrm{CR}-1/\kappa}.
\label{eq:saturation}
\end{equation}
At $\mathrm{CR}=0.4$ and $\kappa=2.5$, this threshold is $M^\star=5d$. Below saturation, wider groups can support larger dictionaries. Beyond saturation, the dictionary \rev{has} a fixed \rev{storage cost of} $d^2$ values, so the nonzero coefficients per output column obey $s=(1-\mathrm{CR})d-d^2/M$ and approach $(1-\mathrm{CR})d$ as $M\to\infty$.

At $\mathrm{CR}=0.4$, the allocation gives approximately $\nnz=0.17d$ for a single square projection ($M=d$), $0.27d$ for two such projections, and $0.56d$ for group width $M=25d$ (Fig.~\ref{fig:budget}). These values quantify the gain in per-column coefficient capacity from amortizing the dictionary, without implying that every choice of group membership improves reconstruction accuracy.

\subsection{Interpreting the Group-Size Capacity Control}
\label{app:capacity-control}
The controlled experiment in Table~\ref{tab:capacity-control} uses the dictionary width
$\natoms_8$ and code sparsity $\nnz_8$ allocated at $G=8$ in a given projection pool,
but fits shared dictionaries across $G=16$ members. For a resulting group of total
output width $M_{16}$, the retained-value fraction is
\begin{equation}
\frac{d\natoms_8+\nnz_8 M_{16}}{dM_{16}}
=\frac{\natoms_8}{M_{16}}+\frac{\nnz_8}{d}.
\label{eq:capacity-control-ratio}
\end{equation}
The restricted setting has nominal CR approximately $0.47$, rather than $0.40$. Moving from standard $G=8$ to this setting changes both membership and storage, so it does not isolate a pure group-size effect. Within the $G=16$ partition, restoring the full allocation improves \textit{\rev{Avg7}} by 5.15/4.71 points on Llama/Qwen. Together with the allocation equations, this supports the role of representation capacity in the gains from broader sharing. Table~\ref{tab:budget} provides the per-pool allocations and additional restricted-budget controls.

\begin{table}[!tb]
\caption{Budget per group at CR $0.4$. (a) Mean \rev{dictionary width} $\natoms$ and \rev{support size} $\nnz$ \rev{for consecutive Llama-3-8B} groups. \rev{At} $\revmath{G}=\revmath{4}$, \rev{8}, and \rev{16}, mean \rev{group widths are} $6.3d$, $12.7d$, \rev{and} $25.3d$ \rev{in the} $\revmath{d}=\revmath{4096}$ \rev{pool}. \rev{The corresponding widths in the} \pj{\rev{down}} \rev{pool} ($\revmath{d}=\revmath{14336}$) \rev{are} $1.1d$, $2.3d$, \rev{and} $4.6d$. At $G=4$, a third of the groups \rev{have} $M=2.5d$ and $\natoms=0.75d$. The lower rows \rev{restrict} groups of 16 \rev{to allocations} $(\natoms,\nnz)$ \rev{inherited from} smaller \rev{groups}, in both pools or \rev{only} the $d=4096$ pool, \rev{reducing stored} values. (b) Width range, $\nnz$, $\nnz$ of \pj{k} and \pj{v}, and calibration error $\times10^{2}$ of the narrowest and widest third of groups in the $d=4096$ pool of Llama-3-8B. \rev{Bold marks the highest average accuracy in each panel}. \rev{Allocation controls in} (\rev{a}) \rev{have different realized budgets}.}
\label{tab:budget}
\centering
\scriptsize
\setlength{\tabcolsep}{1.4pt}
\tablecolors
\begin{tabularx}{\columnwidth}{@{\hspace{1pt}}lYYYYYYYY@{\hspace{1pt}}}
\multicolumn{9}{@{\hspace{1pt}}l}{(a) Consecutive groups and the capacity test} \\
\toprule
\rowband & \multicolumn{2}{c}{$d=4096$} & \multicolumn{2}{c}{\pj{down}} & \multicolumn{2}{c}{Llama-3-8B} & \multicolumn{2}{c}{Qwen3.5-9B} \\
\cmidrule(lr){2-3}\cmidrule(lr){4-5}\cmidrule(lr){6-7}\cmidrule(lr){8-9}
\rowband Groups & $\natoms/d$ & $\nnz/d$ & $\natoms/d$ & $\nnz/d$ & CR & \textit{\rev{Avg7}}$\uparrow$ & CR & \textit{\rev{Avg7}}$\uparrow$ \\
\midrule
$G=4$ & 0.92 & 0.42 & 0.47 & 0.19 & 0.40 & 62.66 & 0.40 & 64.81 \\
$G=8$ & 1.00 & 0.52 & 0.72 & 0.29 & 0.40 & 63.93 & 0.40 & 68.83 \\
$G=16$ & 1.00 & 0.56 & 0.97 & 0.39 & 0.40 & \textbf{66.61} & 0.40 & \textbf{70.58} \\
\midrule
\multicolumn{9}{@{\hspace{1pt}}l}{Groups of 16 with the $(\natoms,\nnz)$ of} \\
$G=8$, both pools & 1.00 & 0.52 & 0.72 & 0.29 & 0.47 & 61.46 & 0.47 & 65.87 \\
$G=8$, $d=4096$ only & 1.00 & 0.52 & 0.97 & 0.39 & 0.43 & 66.19 & 0.43 & 69.47 \\
$G=4$, both pools & 1.00 & 0.44 & 0.47 & 0.19 & 0.57 & 47.84 & 0.56 & 50.49 \\
$G=4$, $d=4096$ only & 1.00 & 0.44 & 0.97 & 0.39 & 0.49 & 63.10 & 0.49 & 65.09 \\
\bottomrule
\end{tabularx}

\medskip
\tablecolors
\begin{tabularx}{\columnwidth}{@{\hspace{1pt}}lcYYcc@{\hspace{1pt}}}
\multicolumn{6}{@{\hspace{1pt}}l}{(b) Partition rules at $G=16$, Llama-3-8B} \\
\toprule
\rowband Grouping & $M/d$ & $\nnz/d$ & $\nnz_{\pj{kv}}/d$ & Err.~narrow/wide & \textit{\rev{Avg7}}$\uparrow$ \\
\midrule
Consecutive & 21.5--27.3 & 0.56 & 0.56 & 0.64~/~0.84 & 66.61 \\
Random & 16.5--34.8 & 0.56 & 0.56 & 0.64~/~0.85 & \sd{66.26}{0.30} \\
Projection & 4.0--56.0 & 0.50 & 0.37 & 1.28~/~1.09 & 65.85 \\
HB & 9.3--35.5 & 0.55 & 0.53 & 1.23~/~0.69 & 66.27 \\
Consec., \pj{v}, \pj{o} isolated & 10.0--33.0 & 0.55 & 0.53 & 1.23~/~0.70 & 66.41 \\
HB + width balance & 23.0--26.8 & 0.56 & 0.56 & 0.86~/~0.67 & \textbf{66.76} \\
\bottomrule
\end{tabularx}

\end{table}

\subsection{Activation Geometry and Coefficient Representation}
\label{app:dict}
The ablations in main Table~\ref{tab:factorization-summary} examine how the activation metric and coefficient representation affect compression quality. Applying one activation metric to an entire group, or to an entire input-dimension pool, lowers accuracy relative to retaining each member's metric. Dense coefficients also reduce accuracy at the same nominal budget. Because they require a narrower dictionary, this comparison measures the combined effect of coefficient sparsity and dictionary capacity.

The frozen-dictionary control in main Table~\ref{tab:factorization-summary} retains the SVD initialization while continuing to fit coefficients. Updating the dictionary improves accuracy on both models, supporting the alternating optimization used by GeoGroup.

\section{Group Construction and Composition}
\label{app:grouping-all}
The experiments in Section~\ref{sec:exp-grouping} show that group composition affects both reconstruction compatibility and the budget available to each projection. We provide the construction rules, examine consecutive grouping choices, and use width-aware alternatives to test the interaction between distance and capacity.

\subsection{Grouping Algorithms}
\label{app:grouping-construction}

Only projections with equal input dimension $d$ share a dictionary. Their output widths $m_i$ may differ. Within each such pool, we construct groups of target size $G$ using the following rules.

\paragraph{Random.}
We draw a uniform random partition of each input-dimension pool. The main results average over multiple grouping seeds, as specified in the experimental setup.

\paragraph{Hierarchical Blossom (HB).}
For each projection, define its whitened weights $\mathbf{Z}_i=\mathbf{L}_i\mathbf{W}_i$. Following \geopair{}~\citep{geopair}, for $m_i<m_j$ we compute the relative best-window distance
\begin{equation}
\delta_{ij}=\min_{0\le t\le m_j-m_i}
\frac{\|\mathbf{Z}_i-\mathbf{Z}_j[:,t:t+m_i]\|_F}
{\|\mathbf{Z}_i\|_F+\epsilon}.
\label{eq:distance}
\end{equation}
Here $\epsilon>0$ stabilizes the denominator. Minimum-cost maximum-cardinality Blossom matching~\citep{edmonds1965paths} first forms projection pairs. At subsequent levels, two groups $\mathcal S_a,\mathcal S_b$ are compared by their mean cross-group member distance,
\begin{equation}
\delta(\mathcal S_a,\mathcal S_b)
=\frac{1}{|\mathcal S_a|\,|\mathcal S_b|}
\sum_{i\in\mathcal S_a}\sum_{j\in\mathcal S_b}\delta_{ij},
\label{eq:group-distance}
\end{equation}
and matching is repeated until the desired group size is reached. Distances are calculated only for compatible input dimensions. When output widths differ, the smaller width is aligned with the best window of the larger matrix.

\paragraph{Consecutive.}
Projections are ordered by Transformer layer and a fixed within-layer order of projection types, then partitioned into consecutive groups of size $G$. The default \emph{Mixed} variant combines attention and MLP projections. Additional variants restrict pools by projection type (\emph{Projection}), common input activations (\emph{Input family}), or attention versus MLP block (\emph{Block family}). Unless otherwise stated, \rev{remaining projections} are \rev{factorized individually}. This rule uses neither calibration statistics nor similarity search.

\paragraph{Why width matters.}
A group shares one dictionary but stores the coefficients of all $M=\sum_{i\in\mathcal S}m_i$ output columns. Consequently, even groups with the same $G$ may have different $M$ and different sparsity budgets under Eq.~\eqref{eq:support-budget}. This coupling between membership and storage allocation motivates the width-balance analyses in Appendix~\ref{app:grouping}.

\subsection{Consecutive Grouping and Remainder Handling}
\label{app:consecutive}
Table~\ref{tab:schemes} varies projection-type restrictions, layer ordering, and per-type budgets. On Llama, restricting groups to individual projection types reduces the support of narrow key and value projections from $0.56d$ to $0.37d$ at $G=16$. Restoring the Mixed per-type budget recovers 66.58 \textit{\rev{Avg7}}, close to the Mixed result of 66.61. The projection-level errors in Table~\ref{tab:app-types-l8} support the same explanation: group restrictions affect accuracy partly through the capacity assigned to each type.

Ordering and projection restrictions offer additional tuning opportunities. On Qwen, Projection grouping with the Mixed per-type budget reaches 71.13 \textit{\rev{Avg7}}. This comparison merges each pool's remainder into one smaller group, giving a Mixed reference of 70.51 rather than the main setting's 70.58. The alternatives therefore refine the choice of grouping and allocation without changing the rationale for a default that requires no similarity search.
\begin{table}[!tb]
\caption{Design choices of Consecutive grouping at CR $=0.4$ (\rev{average accuracy} \%). $\nnz_{\pj{kv}}/d$ is the number of nonzeros per column of the \pj{k} and \pj{v} projections on Llama-3-8B at $G=16$. On Qwen3.5-9B, every variant merges the remainder of each pool into one smaller group, so Mixed scores 70.51 here and 70.58 in Table~\ref{tab:main}. Bold marks the highest \rev{average accuracy} in each column.}
\label{tab:schemes}
\centering
\footnotesize
\setlength{\tabcolsep}{2pt}
\tablecolors
\begin{tabularx}{\columnwidth}{@{\hspace{1pt}}lYYYc@{\hspace{1pt}}}
\toprule
\rowband & & \multicolumn{2}{c}{Llama-3-8B} & Qwen3.5-9B \\
\cmidrule(lr){3-4}\cmidrule(lr){5-5}
\rowband Consecutive variant & $\nnz_{\pj{kv}}/d$ & $G{=}4$ & $G{=}16$ & $G{=}16$ \\
\midrule
\multicolumn{5}{@{\hspace{3pt}}l}{\textit{Which types share a group}} \\
\quad Mixed (default) & 0.56 & 62.66 & 66.61 & 70.51 \\
\quad Projection & 0.37 & 61.94 & 65.85 & 70.57 \\
\quad Input family & 0.47 & 62.80 & \textbf{66.92} & 70.92 \\
\multicolumn{5}{@{\hspace{3pt}}l}{\textit{Budget of the narrow \pj{k}, \pj{v}}} \\
\quad Proj.\ + Mixed budget & 0.56 & \textbf{63.34} & 66.58 & \textbf{71.13} \\
\quad \pj{k}, \pj{v} in own groups & 0.37 & 61.09 & 65.77 & 70.12 \\
\quad \pj{k} with \pj{v} & 0.37 & 63.18 & 66.89 & 70.47 \\
\multicolumn{5}{@{\hspace{3pt}}l}{\textit{Layer order}} \\
\quad Mixed, every other layer & 0.56 & 63.12 & 66.60 & 69.88 \\
\quad Projection, shuffled & 0.37 & 61.03 & 65.47 & 69.74 \\
\bottomrule
\end{tabularx}

\end{table}

\begin{table}[!tb]
\caption{Budget and error by projection type (Llama-3-8B, $G=16$, CR $0.4$). $\nnz/d$ is the number of nonzeros per column relative to the input dimension, and err.\ denotes the relative weight error $\|\mathbf{W}-\mathbf{D}\mathbf{C}\|_F/\|\mathbf{W}\|_F$. Projection grouping lowers the budget of \pj{k} and \pj{v} and raises their error (\pj{v} from 0.43 to 0.82). Restoring the Mixed budget returns errors and average accuracy close to the Mixed baseline. Bold marks the lowest relative error for each projection type and the highest average accuracy.}
\label{tab:app-types-l8}
\centering
\footnotesize
\setlength{\tabcolsep}{2pt}
\tablecolors
\begin{tabularx}{\columnwidth}{@{\hspace{3pt}}lYYYYYY@{\hspace{3pt}}}
\toprule
\rowband & \multicolumn{2}{c}{Mixed} & \multicolumn{2}{c}{Projection} & \multicolumn{2}{c}{+ Mixed budget} \\
\cmidrule(lr){2-3}\cmidrule(lr){4-5}\cmidrule(lr){6-7}
\rowband Type & $\nnz/d$ & err. & $\nnz/d$ & err. & $\nnz/d$ & err. \\
\midrule
\pj{q} & 0.56 & 0.61 & 0.54 & 0.56 & 0.56 & \textbf{0.55} \\
\pj{k} & 0.56 & 0.47 & 0.37 & 0.62 & 0.56 & \textbf{0.46} \\
\pj{v} & 0.56 & \textbf{0.43} & 0.37 & 0.82 & 0.56 & \textbf{0.43} \\
\pj{o} & 0.56 & \textbf{0.38} & 0.54 & 0.42 & 0.56 & 0.39 \\
gate & 0.56 & 0.34 & 0.58 & \textbf{0.28} & 0.56 & 0.29 \\
up & 0.56 & 0.30 & 0.58 & \textbf{0.26} & 0.56 & 0.27 \\
down & 0.39 & \textbf{0.55} & 0.39 & \textbf{0.55} & 0.39 & \textbf{0.55} \\
\midrule
\textit{Avg7} & \multicolumn{2}{c}{\textbf{66.61}} & \multicolumn{2}{c}{65.85} & \multicolumn{2}{c}{66.58} \\
\bottomrule
\end{tabularx}

\end{table}

When a pool is not divisible by $G$, its remainder may be factorized as singletons, merged into one smaller group, or left dense (Table~\ref{tab:app-tails}). The first two retain the nominal projection budget, while leaving it dense changes the compression level. On Qwen, merging the Input-family remainder raises \textit{\rev{Avg7}} from 70.15 to 70.92. Remainder policies must therefore be held fixed when comparing grouping rules.
\begin{table}[!tb]
\caption{\rev{Remainder handling} when a pool \rev{size is} not \rev{divisible} by $G$ (Qwen3.5-9B, $G=16$, nominal CR $0.4$). Model CR is the share of all weights removed. Merging the remainder into one smaller group \rev{preserves} the model CR \rev{of singleton handling} and \rev{achieves accuracy} within 0.3 points of leaving \rev{the remainder} dense. \rev{Bold marks the highest average accuracy within each partition family}. \rev{Model CR differs across remainder policies}.}
\label{tab:app-tails}
\centering
\footnotesize
\setlength{\tabcolsep}{2pt}
\tablecolors
\begin{tabularx}{\columnwidth}{@{\hspace{3pt}}lYY@{\hspace{3pt}}}
\toprule
\rowband Remainder & Model~CR & \textit{\rev{Avg7}}$\uparrow$ \\
\midrule
\multicolumn{3}{@{\hspace{3pt}}l}{\textit{Mixed}} \\
\quad singletons & 30.92\% & \textbf{70.58} \\
\quad one smaller group & 30.92\% & 70.51 \\
\quad kept dense & 29.76\% & 70.30 \\
\multicolumn{3}{@{\hspace{3pt}}l}{\textit{Input family}} \\
\quad singletons & 30.92\% & 70.15 \\
\quad one smaller group & 30.92\% & 70.92 \\
\quad kept dense & 27.74\% & \textbf{71.09} \\
\bottomrule
\end{tabularx}

\end{table}

\subsection{Distance, Width, and Representation Capacity}
\label{app:grouping}
The best-window distance in Eq.~\eqref{eq:distance} gives narrow projections many candidate alignments within a wider matrix. In the Llama $d=4096$ pool, every gate and up projection has a key projection as its nearest neighbor (Table~\ref{tab:app-nn}). This pattern helps explain why hierarchical matching can form groups with uneven widths. Narrower groups devote a larger share of their budget to the dictionary and retain fewer coefficients per output column.

Table~\ref{tab:hb-fix} tests interventions on this interaction. Width balancing improves Llama \textit{\rev{Avg7}} from 66.27 to 66.76 despite increasing mean within-group distance from 1.44 to 3.72. Conversely, local search lowers distance to 1.39 but gives 66.23 \textit{\rev{Avg7}}. These controls show why distance alone is insufficient to predict compression quality. The interventions are defined as follows:
\begin{itemize}[leftmargin=*,itemsep=2pt,topsep=3pt,parsep=0pt]
\item \textit{Width-aware matching.} A merge cost is multiplied by $\exp(\beta(\nnz_{\max}-\nnz_{AB})/d)$, where $\nnz_{AB}$ is the merged group's support and $\nnz_{\max}$ the largest support among candidate merges. We use $\beta=7.6$, fitted to calibration error as a function of $\nnz/d$. The width balance variant instead swaps compatible projections after matching until group widths lie within 10\% of the pool mean, minimizing the number of swaps.
\item \textit{Input families and equalized budgets.} The first variant starts matching from query/key/value and gate/up families rather than individual projections. The second retains the HB partition and assigns every group in a pool the largest common support size that respects the pool budget.
\item \textit{Distance-focused alternatives.} Local search swaps equal-shape projections when this reduces mean within-group distance, so group widths stay fixed. Balanced $k$-medoids alternates medoid updates with assignments of exactly $G$ projections per cluster, and its width balance variant adds the swaps defined above.
\item \textit{Budget-aware matching.} Candidate merges use predicted calibration error $|g|(\tau+(1-\tau)e^{-\beta\nnz/d})$, where $\tau$ is the whitened energy outside the leading $\natoms$ singular directions of the summed member Gram matrix.
\item \textit{Balanced spectral clustering.} Distances define affinities $\exp(-\delta^2/(2\sigma^2))$, with $\sigma$ the median pool distance. Leading eigenvectors of the normalized affinity matrix embed the projections, followed by $k$-means constrained to $G$ members per cluster~\citep{ng2001spectral}.
\end{itemize}
Balanced spectral clustering reaches 67.10 \textit{\rev{Avg7}} on Llama, and equalized-budget HB reaches 71.03 on Qwen. Other interventions yield smaller gains or losses. Together, the results support capacity-aware grouping while establishing Consecutive as a competitive choice without distance computation or partition search.
\begin{table}[!tb]
\caption{Type of each projection's nearest neighbour under the \geopair{} distance (Llama-3-8B, $d=4096$ pool, 32 layers). Rows give the projection and columns its nearest neighbour. No projection is closest to its own type, and every gate and up projection is closest to a \pj{k} projection, which collects half of all nearest-neighbour links.}
\label{tab:app-nn}
\centering
\footnotesize
\setlength{\tabcolsep}{2pt}
\tablecolors
\begin{tabularx}{\columnwidth}{@{\hspace{3pt}}lYYYYYY@{\hspace{3pt}}}
\toprule
\rowband Projection & \pj{q} & \pj{k} & \pj{v} & \pj{o} & gate & up \\
\midrule
\pj{q} & 0 & 0 & 0 & \textbf{29} & 0 & 3 \\
\pj{k} & 0 & 0 & 0 & \textbf{18} & 8 & 6 \\
\pj{v} & 0 & 2 & 0 & \textbf{30} & 0 & 0 \\
\pj{o} & 0 & \textbf{30} & 2 & 0 & 0 & 0 \\
gate & 0 & \textbf{32} & 0 & 0 & 0 & 0 \\
up & 0 & \textbf{32} & 0 & 0 & 0 & 0 \\
\midrule
total & 0 & 96 & 2 & 77 & 8 & 9 \\
\bottomrule
\end{tabularx}

\end{table}

\begin{table}[!tb]
\caption{Width- and distance-aware grouping variants at CR $0.4$ and $G=16$ (\rev{average accuracy} \%). $\bar\delta$ is the mean \geopair{} distance between members of the same group on Llama-3-8B. Balancing group widths \rev{increases} HB \rev{accuracy} to 66.6--66.8 while raising $\bar\delta$ from 1.44 to 2.9--3.7, and local search lowers $\bar\delta$ to 1.39 without \rev{improving} accuracy. On Qwen3.5-9B, width-aware variants raise HB from 69.88 to 70.07--70.53, equalized budgets \rev{increase it} to 71.03, and balanced spectral clustering reaches 70.84, against 70.58 for Consecutive. Bold marks the highest \rev{average accuracy} in each column.}
\label{tab:hb-fix}
\centering
\scriptsize
\setlength{\tabcolsep}{2pt}
\tablecolors
\begin{tabularx}{\columnwidth}{@{\hspace{1pt}}lYcYY@{\hspace{1pt}}}
\toprule
\rowband & \multicolumn{2}{c}{Llama-3-8B} & Qwen3.5-9B & Llama-3.2-1B \\
\cmidrule(lr){2-3}
\rowband Grouping & \textit{\rev{Avg7}}$\uparrow$ & $\bar\delta$ & \textit{\rev{Avg7}}$\uparrow$ & \textit{\rev{Avg7}}$\uparrow$ \\
\midrule
HB & 66.27 & 1.44 & 69.88 & 52.18 \\
\quad + width-aware matching & 66.61 & 2.88 & 70.07 & 51.95 \\
\quad + width balance & 66.76 & 3.72 & 70.53 & 51.44 \\
\quad + input families & 66.71 & 3.60 & 69.93 & 51.60 \\
\quad + equalized budgets & 66.43 & 1.44 & \textbf{71.03} & \textbf{52.53} \\
\quad + local search & 66.23 & 1.39 & 70.47 & 51.10 \\
Budget-aware matching & 66.23 & 4.64 & 69.93 & 51.30 \\
Balanced $k$-medoids & 66.80 & 1.74 & 69.65 & 51.57 \\
\quad + width balance & 66.29 & 2.72 & 69.42 & 51.53 \\
Balanced spectral clustering & \textbf{67.10} & 2.44 & 70.84 & 51.06 \\
\midrule
Consecutive (default) & 66.61 & 3.35 & 70.58 & 51.45 \\
\bottomrule
\end{tabularx}

\end{table}

\section{Extended Language-Model Evaluation}
\label{app:further}
The following results expand the main quality comparisons across tasks, group sizes, compression levels, and model architectures. \rev{CR applies to} the \rev{compressed projection subset}, \rev{with} selective \rev{scopes identified explicitly}.

\subsection{Task-Level Comparisons and Group-Size Sweeps}
\label{app:full}
Table~\ref{tab:main} reports the task-level scores and LAMBADA, WikiText-2, and C4 perplexities underlying the main comparison. Table~\ref{tab:app-sweep-all} extends the group-size and compression sweep to Llama-3.2-1B. The $G=2$ HB rows use Blossom pairs with Jacobi PCG and 30 rounds. They are distinct from the analytic GeoPair baseline, which uses 50 rounds. Random results include variation over grouping seeds where multiple runs are available.

\paragraph{Aggressive compression.}
At CR 0.6, consecutive $G=16$ achieves 47.4 \textit{\rev{Avg7}} on Llama, compared with 36.0 for GeoPair (Table~\ref{tab:app-extreme}). At CR 0.8, all evaluated methods fall within 27.8--28.5. These settings delimit the compression range over which the quality advantages in the main paper are useful.
\label{app:extreme}
\begin{table}[!tb]
\caption{Extreme compression on Llama-3-8B. At CR $0.6$, Consecutive $G=16$ scores 11.4 points above \geopair{}. At CR $0.8$, every method \rev{achieves average accuracy of} 27.8--28.5\%. PPL is LAMBADA perplexity. \rev{Bold} marks the highest \rev{average accuracy and lowest PPL within each CR}.}
\label{tab:app-extreme}
\centering
\footnotesize
\setlength{\tabcolsep}{2pt}
\tablecolors
\begin{tabularx}{\columnwidth}{@{\hspace{3pt}}lYYYY@{\hspace{3pt}}}
\toprule
\rowband & \multicolumn{2}{c}{CR $0.6$} & \multicolumn{2}{c}{CR $0.8$} \\
\cmidrule(lr){2-3}\cmidrule(lr){4-5}
\rowband Method & \textit{\rev{Avg7}}$\uparrow$ & PPL$\downarrow$ & \textit{\rev{Avg7}}$\uparrow$ & PPL$\downarrow$ \\
\midrule
\rowgeopair \geopair{} & 36.01 & 1.7e3 & 27.76 & 1.6e6 \\
\rowoursalt Consecutive $G{=}2$ & 40.20 & 752.8 & 28.17 & \textbf{8.9e5} \\
\rowoursalt Consecutive $G{=}4$ & 41.33 & 272.9 & 28.33 & 2.5e6 \\
\rowoursalt Consecutive $G{=}8$ & 45.33 & 65.17 & 28.17 & 2.4e6 \\
\rowours Consecutive $G{=}16$ & 47.38 & 35.83 & \textbf{28.48} & 1.4e6 \\
\rowoursalt Input family $G{=}16$ & \textbf{49.04} & \textbf{22.13} & 28.41 & 1.7e6 \\
\bottomrule
\end{tabularx}

\end{table}

\begin{table*}[!t]
\caption{Comparison with factorization baselines (accuracy in \%). LAMB., Wiki2, and C4 denote perplexity on LAMBADA, WikiText-2, and C4, respectively. PPL at CR 0.5 denotes LAMBADA perplexity. Random is the mean $\pm$ SD over 3--7 grouping seeds, and Basis Sharing uses $G=2$. GeoGroup rows use HB, Random, or Consecutive grouping. ROCKET~\citep{ali2026rocket} and COMPOT~\citep{makhov2026compot} are rerun under the same conditions as GeoGroup and GeoPair. Bold marks the best compressed value within each model.}
\label{tab:main}
\centering
\scriptsize
\setlength{\tabcolsep}{1.4pt}
\renewcommand{\arraystretch}{1.1}
\tablecolors
\begin{tabularx}{\textwidth}{@{\hspace{3pt}}lY*{7}{Y}YYYY@{\hspace{6pt}}YY@{\hspace{3pt}}}
\toprule
\rowband & & \multicolumn{11}{c}{CR $=0.4$} & \multicolumn{2}{c}{CR $=0.5$} \\
\cmidrule(lr){3-13}\cmidrule(lr){14-15}
\rowband Method & $G$ & ARC-C & ARC-E & HellaS. & MMLU & PIQA & RACE & SciQ & \textit{Avg7}$\uparrow$ & LAMB.$\downarrow$ & Wiki2$\downarrow$ & C4$\downarrow$ & \textit{Avg7}$\uparrow$ & PPL$\downarrow$ \\
\midrule
\model{Llama-3-8B}, dense & -- & 54.0 & 77.6 & 79.3 & 62.1 & 80.9 & 40.1 & 94.5 & 69.78 & 3.13 & 6.14 & 9.47 & 69.78 & 3.13 \\
SVD-LLM & 1 & 23.5 & 31.9 & 32.9 & 23.0 & 58.7 & 25.3 & 41.6 & 33.82 & 3.2e3 & 396.8 & 185.2 & 28.75 & 7.3e4 \\
ROCKET & 1 & 34.0 & 62.3 & 57.1 & 37.5 & 72.1 & 37.9 & 87.6 & 55.51 & 10.04 & 46.05 & 33.92 & 44.43 & 164.6 \\
COMPOT & 1 & 33.2 & 55.9 & 56.0 & 30.4 & 71.1 & 36.3 & 86.0 & 52.71 & 21.35 & 54.36 & 35.42 & 40.67 & 530.6 \\
Basis Sharing & 2 & 23.8 & 33.5 & 34.1 & 23.0 & 59.9 & 26.8 & 44.2 & 35.04 & 2.9e3 & 421.7 & 195.2 & 29.59 & 7.1e4 \\
\rowgeopair \geopair{} & 2 & 39.0 & 68.4 & 65.0 & 44.5 & 74.5 & 40.1 & 92.1 & 60.50 & 5.21 & 18.35 & 19.73 & 48.65 & 23.42 \\ \cmidrule(lr){1-15}
\rowoursalt \method{} (HB) & 16 & 48.9 & 74.8 & \textbf{74.1} & 53.7 & \textbf{78.7} & \textbf{41.0} & 92.8 & 66.27 & 3.10 & \textbf{8.68} & \textbf{12.88} & 60.72 & 3.99 \\
\rowoursalt \method{} (Random) & 16 & \stk{48.2}{1.0} & \stk{75.6}{0.9} & \stk{73.3}{0.2} & \stk{54.1}{0.4} & \stk{78.3}{0.2} & \stk{40.7}{0.4} & \stk{\textbf{93.6}}{0.3} & \stk{66.26}{0.30} & \stk{2.86}{0.04} & \stk{9.06}{0.11} & \stk{13.30}{0.10} & \stk{60.85}{0.51} & \stk{3.86}{0.14} \\
\rowours \method{} (Consecutive) & 16 & \textbf{49.3} & \textbf{76.1} & 73.8 & \textbf{54.8} & \textbf{78.7} & 40.1 & \textbf{93.6} & \textbf{66.61} & \textbf{2.78} & 9.07 & 13.23 & \textbf{61.09} & \textbf{3.83} \\
\midrule
\model{Qwen3.5-9B}, dense & -- & 55.9 & 74.3 & 78.1 & 78.7 & 80.0 & 43.0 & 95.9 & 72.26 & 3.88 & 8.65 & 12.23 & 72.26 & 3.88 \\
SVD-LLM & 1 & 23.0 & 38.6 & 37.7 & 23.0 & 59.9 & 30.5 & 79.1 & 41.67 & 65.36 & 115.6 & 76.95 & 35.51 & 814.1 \\
ROCKET & 1 & 38.4 & 68.1 & 57.8 & 51.4 & 72.5 & 40.1 & 93.6 & 60.27 & 6.25 & 23.81 & 24.39 & 48.87 & 15.76 \\
COMPOT & 1 & 38.3 & 65.3 & 56.4 & 51.9 & 71.1 & 38.9 & 92.2 & 59.17 & 5.62 & 26.97 & 24.92 & 47.53 & 14.81 \\
Basis Sharing & 2 & 25.7 & 41.7 & 41.1 & 28.0 & 61.1 & 34.4 & 84.5 & 45.20 & 21.43 & 59.36 & 39.32 & 38.51 & 86.95 \\
\rowgeopair \geopair{} & 2 & 42.5 & 71.3 & 57.8 & 46.6 & 73.5 & 40.0 & 92.3 & 60.57 & 8.72 & 26.00 & 28.72 & 42.06 & 153.4 \\ \cmidrule(lr){1-15}
\rowoursalt \method{} (HB) & 16 & 54.5 & 79.8 & \textbf{72.8} & 67.6 & 77.9 & 40.5 & 96.2 & 69.88 & 3.43 & \textbf{11.23} & \textbf{16.64} & 62.92 & 5.04 \\
\rowoursalt \method{} (Random) & 16 & \stk{53.5}{1.8} & \stk{79.5}{0.6} & \stk{71.2}{0.5} & \stk{\textbf{68.8}}{0.7} & \stk{\textbf{78.1}}{0.1} & \stk{40.7}{0.6} & \stk{95.6}{0.5} & \stk{69.61}{0.46} & \stk{3.40}{0.08} & \stk{11.25}{0.22} & \stk{16.68}{0.19} & \stk{62.78}{0.81} & \stk{\textbf{4.45}}{0.15} \\
\rowours \method{} (Consecutive) & 16 & \textbf{57.6} & \textbf{81.0} & 71.6 & 68.6 & 77.6 & \textbf{41.2} & \textbf{96.5} & \textbf{70.58} & \textbf{3.31} & 11.59 & 17.05 & \textbf{64.38} & 4.47 \\
\bottomrule
\end{tabularx}

\end{table*}

\begin{table*}[!t]
\caption{\rev{Average accuracy} (\%) \rev{across the evaluated} grouping \rev{rules}, group \rev{sizes}, \rev{models}, and \rev{compression ratios} (available perplexities in Tables~\ref{tab:main}, \ref{tab:app-transfer}, and \ref{tab:app-extreme}). HB at $G=2$ is Blossom matching with Jacobi PCG and 30 rounds, whereas the analytic GeoPair baseline uses 50 rounds. Pair-GEVD controls are listed separately in Table~\ref{tab:precond}. Random is the mean $\pm$ SD over grouping seeds, or a single seed where no SD is shown. \rev{Bold marks the highest average accuracy for each model and CR across all reported groupings and group sizes}.}
\label{tab:app-sweep-all}
\centering
\footnotesize
\setlength{\tabcolsep}{2pt}
\tablecolors
\begin{tabularx}{\textwidth}{@{\hspace{3pt}}lYYYYYYYYY@{\hspace{3pt}}}
\toprule
\rowband & & \multicolumn{3}{c}{Llama-3-8B} & \multicolumn{3}{c}{Qwen3.5-9B} & \multicolumn{2}{c}{Llama-3.2-1B} \\
\cmidrule(lr){3-5}\cmidrule(lr){6-8}\cmidrule(lr){9-10}
\rowband Grouping & $G$ & 0.2 & 0.4 & 0.5 & 0.2 & 0.4 & 0.5 & 0.4 & 0.5 \\
\midrule
\rowgeopair \geopair{} & 2 & 67.62 & 60.50 & 48.65 & 72.15 & 60.57 & 42.06 & 46.22 & 39.84 \\
\midrule
HB & 2 & 67.05 & 59.78 & 48.99 & 71.03 & 57.81 & 39.05 & 45.46 & 39.06 \\
HB & 4 & 68.08 & 62.60 & 54.37 & 72.18 & 64.75 & 43.61 & 47.52 & 42.06 \\
HB & 8 & 68.67 & 63.82 & 58.02 & 72.31 & 68.88 & 60.01 & 49.65 & 44.80 \\
HB & 16 & \textbf{69.59} & 66.27 & 60.72 & \textbf{72.63} & 69.88 & 62.92 & \textbf{52.18} & 46.60 \\
\midrule
Random & 2 & \sd{67.30}{0.11} & \sd{59.12}{0.59} & \sd{47.81}{1.41} & 71.41 & \sd{57.58}{3.74} & \na & \sd{42.98}{1.08} & \sd{34.65}{0.95} \\
Random & 4 & \sd{68.01}{0.19} & \sd{62.27}{0.41} & \sd{54.66}{0.18} & 71.43 & \sd{65.31}{0.99} & \sd{50.42}{4.34} & \sd{48.56}{0.04} & \sd{41.99}{0.56} \\
Random & 8 & 68.94 & \sd{64.51}{0.41} & \sd{58.29}{0.12} & 72.13 & \sd{68.29}{0.73} & \sd{60.01}{1.20} & \sd{50.20}{0.39} & \sd{44.85}{0.12} \\
Random & 16 & \sd{69.58}{0.30} & \sd{66.26}{0.30} & \sd{60.85}{0.51} & \sd{72.37}{0.49} & \sd{69.61}{0.46} & \sd{62.78}{0.81} & \sd{51.76}{0.09} & \sd{\textbf{47.08}}{0.28} \\
\midrule
Consecutive & 2 & 67.62 & 61.36 & 52.60 & 70.46 & 61.49 & 49.75 & 46.28 & 40.83 \\
Consecutive & 4 & 67.80 & 62.66 & 54.30 & 71.67 & 64.81 & 51.97 & 48.46 & 42.89 \\
Consecutive & 8 & 68.81 & 63.93 & 59.19 & 72.61 & 68.83 & 60.01 & 49.77 & 45.83 \\
\rowours Consecutive & 16 & 69.30 & \textbf{66.61} & \textbf{61.09} & 72.32 & \textbf{70.58} & \textbf{64.38} & 51.45 & 46.78 \\
\bottomrule
\end{tabularx}

\end{table*}

\subsection{Transfer Across Model Architectures}
\label{app:transfer}
Table~\ref{tab:app-transfer} gives complete results for Llama-3.2-1B, Phi-4, and Gemma-3-4B. Consecutive $G=16$ improves \textit{\rev{Avg7}} over GeoPair on all three models. The gains are distributed unevenly across tasks: Llama-3.2-1B improves most on ARC-Easy and HellaSwag, while Phi-4 exceeds its dense reference on some tasks and remains below it on others. Gemma-3-4B uses the instruction-tuned checkpoint, whose dense HellaSwag score is lower than those of all GeoGroup variants. These task-level results complement the aggregate transfer gains without implying uniform improvement over the dense models.
\begin{table*}[!t]
\caption{\rev{Per-task results across additional} models at CR $0.4$ (accuracy \%, LAMBADA PPL). Blossom + PCG factorizes the \geopair{} pairs with our PCG solver. Random is the mean $\pm$ SD over three grouping seeds. Gemma-3-4B is the instruction-tuned model, and its dense HellaSwag score lies below every variant of \method{}. \rev{Bold marks the highest accuracy and lowest perplexity among compressed configurations within each model}.}
\label{tab:app-transfer}
\centering
\footnotesize
\setlength{\tabcolsep}{2pt}
\tablecolors
\begin{tabularx}{\textwidth}{@{\hspace{3pt}}lYYYYYYYYYY@{\hspace{3pt}}}
\toprule
\rowband Method & $G$ & ARC-C & ARC-E & HellaS. & MMLU & PIQA & RACE & SciQ & \textit{\rev{Avg7}}$\uparrow$ & PPL$\downarrow$ \\
\midrule
\model{Llama-3.2-1B}, dense & \dense{--} & \dense{37.0} & \dense{61.7} & \dense{64.3} & \dense{37.7} & \dense{74.8} & \dense{37.4} & \dense{91.1} & \dense{57.71} & \dense{5.43} \\
Basis Sharing & 2 & 23.5 & 27.1 & 26.8 & 23.3 & 52.9 & 24.0 & 28.5 & 29.44 & 9.1e4 \\
\rowgeopair \geopair{} & 2 & 26.5 & 45.6 & 45.1 & 26.1 & 67.3 & 33.0 & 79.9 & 46.22 & 21.85 \\
Blossom + PCG & 2 & 26.8 & 46.0 & 45.2 & 26.3 & 67.6 & 32.9 & 79.6 & 46.34 & 22.73 \\
\rowoursalt \method{} (HB) & 16 & 31.0 & 55.5 & \textbf{54.9} & \textbf{29.6} & \textbf{71.3} & \textbf{35.9} & \textbf{87.1} & \textbf{52.18} & 8.01 \\
\rowoursalt \method{} (Random) & 16 & \stk{\textbf{31.7}}{0.6} & \stk{54.9}{1.0} & \stk{54.5}{0.4} & \stk{28.7}{0.2} & \stk{71.0}{0.8} & \stk{35.5}{1.0} & \stk{86.0}{0.4} & \stk{51.76}{0.09} & \stk{7.44}{0.24} \\
\rowours \method{} (Consecutive) & 16 & \textbf{31.7} & \textbf{55.6} & 54.1 & 27.7 & 71.1 & 34.7 & 85.3 & 51.45 & \textbf{7.25} \\
\midrule
\model{Phi-4}, dense & \dense{--} & \dense{56.0} & \dense{74.0} & \dense{80.3} & \dense{77.9} & \dense{81.0} & \dense{42.0} & \dense{95.5} & \dense{72.39} & \dense{3.52} \\
Basis Sharing & 2 & 37.1 & 59.5 & 59.3 & 44.6 & 71.8 & 39.6 & 87.9 & 57.11 & 9.61 \\
\rowgeopair \geopair{} & 2 & 47.6 & 72.3 & 69.8 & 55.4 & 76.0 & 42.2 & 93.2 & 65.23 & 4.38 \\
Blossom + PCG & 2 & 47.6 & 72.4 & 69.8 & 55.3 & 76.1 & 42.1 & 93.3 & 65.24 & 4.37 \\
\rowoursalt \method{} (HB) & 16 & 57.3 & 80.6 & \textbf{76.0} & 68.3 & \textbf{81.0} & 42.0 & 95.3 & 71.50 & 3.47 \\
\rowoursalt \method{} (Random) & 16 & \stk{56.9}{0.9} & \stk{81.4}{0.6} & \stk{75.6}{0.2} & \stk{68.6}{0.3} & \stk{80.3}{0.4} & \stk{41.7}{0.2} & \stk{\textbf{95.4}}{0.5} & \stk{71.40}{0.25} & \stk{3.13}{0.02} \\
\rowours \method{} (Consecutive) & 16 & \textbf{58.5} & \textbf{81.5} & 75.9 & \textbf{70.0} & 80.1 & \textbf{43.1} & \textbf{95.4} & \textbf{72.07} & \textbf{3.01} \\
\midrule
\model{Gemma-3-4B}, dense & \dense{--} & \dense{44.8} & \dense{63.4} & \dense{43.5} & \dense{53.4} & \dense{68.1} & \dense{38.1} & \dense{87.4} & \dense{56.95} & \dense{7.69} \\
Basis Sharing & 2 & 24.3 & 31.7 & 32.6 & 23.0 & 56.3 & 28.0 & 52.1 & 35.43 & 509.0 \\
\rowgeopair \geopair{} & 2 & 31.2 & 43.2 & 41.0 & 25.4 & 65.7 & 34.6 & 72.2 & 44.77 & 24.08 \\
Blossom + PCG & 2 & 31.1 & 45.0 & 40.8 & 27.1 & 65.6 & 35.8 & 74.8 & 45.73 & 26.88 \\
\rowoursalt \method{} (HB) & 16 & 39.4 & \textbf{57.5} & 48.5 & \textbf{45.2} & 68.1 & 39.4 & \textbf{84.0} & 54.59 & 5.97 \\
\rowoursalt \method{} (Random) & 16 & \stk{38.9}{2.3} & \stk{55.0}{2.0} & \stk{49.5}{0.8} & \stk{38.2}{13.4} & \stk{69.5}{0.8} & \stk{\textbf{39.8}}{0.7} & \stk{83.8}{0.9} & \stk{53.54}{2.33} & \stk{6.01}{0.14} \\
\rowours \method{} (Consecutive) & 16 & \textbf{40.4} & 57.1 & \textbf{50.6} & 44.2 & \textbf{70.6} & 38.8 & 82.9 & \textbf{54.95} & \textbf{5.86} \\
\bottomrule
\end{tabularx}

\end{table*}

\subsection{Reasoning and Generation Tasks}
\label{app:bench}
Table~\ref{tab:app-bench} extends the Llama evaluation to WinoGrande~\citep{sakaguchi2020winogrande}, BoolQ~\citep{clark2019boolq}, OpenBookQA~\citep{mihaylov2018openbookqa}, RTE~\citep{wang2019glue}, TruthfulQA~\citep{lin2022truthfulqa}, GSM8K~\citep{cobbe2021gsm8k}, and four Open LLM Leaderboard v2 tasks~\citep{fourrier2024leaderboard}: MMLU-Pro, BBH, MuSR, and GPQA~\citep{wang2024mmlupro,suzgun2023bbh,sprague2024musr,rein2024gpqa}. Consecutive $G=16$ improves over GeoPair on three of the five additional multiple-choice tasks and on GSM8K, while the relative ranking varies on the remaining tasks.

Generation is more sensitive to compression than the seven-task average. At CR 0.4, GSM8K accuracy decreases from 48.4 to 21.0 for consecutive $G=16$, compared with 5.3 for GeoPair. The corresponding HumanEval pass@1 measurements are 37.8 for dense, 7.3 for consecutive $G=16$, and 0.0 for GeoPair. These \rev{results show that accuracy retention varies substantially across tasks}.
\begin{table*}[!t]
\caption{Further benchmarks on Llama-3-8B at CR $0.4$ (\%). The multiple-choice tasks are zero-shot, and TQA is TruthfulQA (MC2). GSM8K is five-shot exact match with greedy decoding. The leaderboard tasks use their LM Evaluation Harness configurations, and each score is the unweighted mean over the subtasks, using normalized accuracy for tasks that support this metric and standard accuracy for the remaining tasks. Bold marks the best compressed \rev{result for each metric}.}
\label{tab:app-bench}
\centering
\footnotesize
\setlength{\tabcolsep}{2pt}
\tablecolors
\begin{tabularx}{\textwidth}{@{\hspace{3pt}}lYYYYYYYcYYY@{\hspace{3pt}}}
\toprule
\rowband & & \multicolumn{5}{c}{Multiple choice} & \multicolumn{1}{c}{Gen.} & \multicolumn{4}{c}{Open LLM Leaderboard v2} \\
\cmidrule(lr){3-7}\cmidrule(lr){8-8}\cmidrule(lr){9-12}
\rowband Method & $G$ & WinoG. & BoolQ & OBQA & RTE & TQA & GSM8K & MMLU-Pro & BBH & MuSR & GPQA \\
\midrule
\dense{Dense} & \dense{--} & \dense{73.4} & \dense{81.9} & \dense{45.4} & \dense{69.7} & \dense{42.9} & \dense{48.4} & \dense{31.7} & \dense{45.6} & \dense{37.1} & \dense{28.3} \\
\midrule
\rowgeopair \geopair{} & 2 & 69.2 & 78.7 & 39.0 & \textbf{68.6} & 40.8 & 5.3 & 18.1 & 36.0 & 39.9 & \textbf{27.1} \\
\rowoursalt \method{} (HB) & 16 & \textbf{72.1} & 81.0 & \textbf{42.6} & 67.1 & \textbf{42.1} & \textbf{24.0} & 23.7 & \textbf{41.0} & \textbf{40.2} & 25.3 \\
\rowours \method{} (Consecutive) & 16 & \textbf{72.1} & \textbf{81.2} & 42.2 & 62.8 & 40.7 & 21.0 & \textbf{25.7} & 40.7 & 37.4 & 25.6 \\
\bottomrule
\end{tabularx}

\end{table*}

\subsection{Selective Compression of the Larger Model}
\label{app:selective}
Table~\ref{tab:selective} evaluates Qwen3.8-27B with only up and gate projections compressed. CR applies to this subset and is not directly comparable to full-backbone compression. At CR 0.6 and 0.7, groups of sixteen improve \textit{\rev{Avg7}} over pairs while increasing LAMBADA perplexity. At CR 0.9, pairs perform better on both measures despite the larger dictionary and support allocated to $G=16$. This study illustrates why both task metrics and compression scope matter when selecting a grouping configuration.
\begin{table*}[!t]
\caption{Selective compression of Qwen3.8-27B: only up/gate projections are compressed. CR is nominal on that subset. \rev{Bold marks the highest average accuracy and lowest PPL at each CR where both group sizes are evaluated}.}
\label{tab:selective}

\centering\small
\tablecolors
\begin{tabular}{cccccc}
\toprule

\rowband CR & $G$ & \textit{\rev{Avg7}} (\%) & LAMBADA PPL & $k$ & $s$\\
\midrule
0.6 & 2 & 73.14 & \textbf{3.06} & 3743 & 1496\\
0.6 & 16 & \textbf{74.04} & 3.31 & 4895 & 1958\\
0.7 & 2 & 67.65 & \textbf{3.82} & 2807 & 1122\\
0.7 & 16 & \textbf{68.58} & 4.55 & 3671 & 1467\\
0.8 & 2 & -- & -- & -- & --\\
0.8 & 16 & 54.04 & -- & -- & --\\
0.9 & 2 & \textbf{32.91} & \textbf{1.5e4} & 935 & 374\\
0.9 & 16 & 30.29 & 3.6e5 & 1223 & 489\\

\bottomrule
\end{tabular}
\end{table*}

\rev{Table}~\ref{tab:selective-quality} \rev{expands the CR 0.6 selective Qwen3.8-27B results to individual tasks}. $\revmath{G}=\revmath{16}$ \rev{scores 0.90} \textit{\rev{Avg7}} \rev{points above} $\revmath{G}=\revmath{2}$ \rev{and 0.33 below dense}. \rev{This aggregate advantage is not uniform}: $\revmath{G}=\revmath{2}$ \rev{has higher SciQ and LAMBADA accuracy and lower LAMBADA perplexity}. \rev{Accuracy and latency are evaluated on the same gate}/\rev{up-compressed checkpoints}. \rev{Together}, \rev{they show a measured quality}--\rev{latency trade-off}: $\revmath{G}=\revmath{16}$ \rev{achieves higher average accuracy}, \rev{while} $\revmath{G}=\revmath{2}$ \rev{has lower latency and weight storage}.
\begin{table*}[!t]
\caption{\rev{Per-task quality for Qwen3.8-27B with gate}/\rev{up-only compression at CR 0.6}. \rev{Average accuracy covers the seven tasks from MMLU through SciQ}. \rev{LAMBADA is excluded from the seven-task mean}. \rev{Latency for the same checkpoints is reported in Table}~\ref{tab:decode-qwen27}. \rev{Bold marks the better compressed result}.}
\label{tab:selective-quality}
\centering\small\tablecolors
\begin{tabular}{lrrrrrrrrrr}
\toprule
\rowband & & \multicolumn{7}{c}{\rev{Task accuracy} (\%)} & \multicolumn{2}{c}{\rev{LAMBADA}}\\
\rowband \rev{Method} & \textit{\rev{Avg7}} & \rev{MMLU} & \rev{ARC-E} & \rev{ARC-C} & \rev{HellaSwag} & \rev{PIQA} & \rev{RACE} & \rev{SciQ} & \rev{PPL} & \rev{Acc}.\\
\midrule
\rev{Dense} & \rev{74.37} & \rev{83.5} & \rev{73.1} & \rev{58.6} & \rev{82.8} & \rev{81.7} & \rev{47.6} & \rev{93.3} & \rev{3.52} & \rev{72.1}\\
\rev{G2} & \rev{73.14} & \rev{77.2} & \rev{82.1} & \rev{57.8} & \rev{75.8} & \rev{79.5} & \rev{43.3} & \textbf{\rev{96.2}} & \textbf{\rev{3.06}} & \textbf{\rev{73.6}}\\
\rev{G16} & \textbf{\rev{74.04}} & \textbf{\rev{79.5}} & \textbf{\rev{82.2}} & \textbf{\rev{58.6}} & \textbf{\rev{78.6}} & \textbf{\rev{80.4}} & \textbf{\rev{44.4}} & \rev{94.6} & \rev{3.31} & \rev{72.0}\\
\bottomrule
\end{tabular}
\end{table*}

\subsection{\rev{Selective Compression of DeepSeek-MoE}}
\label{app:moe}
\rev{Table}~\ref{tab:deepseek-moe} \rev{evaluates gate}/\rev{up-only compression of DeepSeek-MoE with HTP and Jacobi-PCG capped at 20 iterations}. \rev{Across CR 0.2}, \rev{0.4}, \rev{and 0.5}, $\revmath{G}=\revmath{64}$ \rev{gives the highest} \textit{\rev{Avg7}}: \rev{63.74}, \rev{63.27}, \rev{and 63.18}, \rev{respectively}, \rev{compared with 63.59 dense}. \rev{At CR 0.5 it exceeds} $\revmath{G}=\revmath{2}$ \rev{by 1.28 points and remains 0.41 below dense}. \rev{At CR 0.2}, \rev{accuracy remains close to dense}. \rev{Perplexity does not follow the same ordering}: $\revmath{G}=\revmath{16}$ \rev{has lower PPL than} $\revmath{G}=\revmath{64}$ \rev{at CR 0.4 and 0.5}.

\begin{table}[t]
\caption{\rev{DeepSeek-MoE quality with gate}/\rev{up-only compression}, \rev{using HTP and Jacobi-PCG with a 20-iteration cap}. \rev{Bold marks the highest average accuracy and lowest perplexity at each CR}.}
\label{tab:deepseek-moe}
\ggtabsetup\tablecolors
\begin{tabular}{ccrr}
\toprule
\rowband \rev{CR} & $\revmath{G}$ & \textit{\rev{Avg7}} (\%) & \rev{PPL}\\
\midrule
\rev{Dense} & -- & \rev{63.59} & \rev{3.29}\\
\rev{0.2} & \rev{2} & \rev{63.09} & \rev{3.20}\\
\rev{0.2} & \rev{16} & \rev{63.45} & \rev{3.19}\\
\rev{0.2} & \rev{64} & \textbf{\rev{63.74}} & \textbf{\rev{3.18}}\\
\rev{0.4} & \rev{2} & \rev{62.52} & \rev{3.11}\\
\rev{0.4} & \rev{16} & \rev{63.20} & \textbf{\rev{3.00}}\\
\rev{0.4} & \rev{64} & \textbf{\rev{63.27}} & \rev{3.04}\\
\rev{0.5} & \rev{2} & \rev{61.90} & \rev{3.09}\\
\rev{0.5} & \rev{16} & \rev{62.73} & \textbf{\rev{2.87}}\\
\rev{0.5} & \rev{64} & \textbf{\rev{63.18}} & \rev{2.90}\\
\bottomrule
\end{tabular}
\end{table}

\section{Multimodal Evaluation}
\label{app:mm}
We evaluate Qwen3.5-9B with a dense vision encoder and compressed language-backbone projections. Mixed calibration uses 128 FineWeb sequences and 128 image--text examples from The Cauldron~\citep{laurencon2024matters}. MM Avg is the unweighted mean accuracy on MMMU~\citep{yue2024mmmu}, MMBench~\citep{liu2024mmbench}, MMMU-Pro~\citep{yue2025mmmupro}, RealWorldQA~\citep{realworldqa2024}, MMStar~\citep{chen2024mmstar}, and HallusionBench~\citep{guan2024hallusionbench}.

\subsection{Compression Quality Across Modalities}
At CR 0.4, HB, Random, and Consecutive groups of sixteen achieve 63.70, 63.58, and 63.60 MM Avg, respectively, compared with 68.48 for dense and 57.33 for GeoPair (Table~\ref{tab:app-mm}). The consistency across grouping rules supports the benefit of sharing beyond pairs in this setting.
\begin{table*}[!t]
\caption{Multimodal accuracy (\%) of Qwen3.5-9B at CR $0.4$ with mixed calibration. \rev{Multimodal average accuracy} is the mean over MMMU, MMBench, MMMU-Pro, RealWorldQA, MMStar and HallusionBench. Random is the mean $\pm$ SD over grouping seeds, and $^\ast$ marks Basis Sharing without its coefficient update. Bold marks the best compressed \rev{result for each metric}.}
\label{tab:app-mm}
\centering
\scriptsize
\setlength{\tabcolsep}{2pt}
\tablecolors
\begin{tabularx}{\textwidth}{@{\hspace{3pt}}lYYYYYYYY@{\hspace{3pt}}}
\toprule
\rowband Method & $G$ & MMMU & MMB. & MMMU-P & RWQA & MMStar & Hall. & MM~Avg$\uparrow$ \\
\midrule
\dense{Dense} & \dense{--} & \dense{55.4} & \dense{90.0} & \dense{53.6} & \dense{75.2} & \dense{65.4} & \dense{71.3} & \dense{68.48} \\
\midrule
SVD-LLM & 1 & 30.8 & 72.3 & 32.7 & 60.1 & 44.5 & 54.0 & 49.06 \\
Basis Sharing$^\ast$ & 16 & 30.6 & 69.2 & 30.8 & 55.3 & 40.7 & 60.0 & 47.77 \\
\rowgeopair \geopair{} & 2 & 40.6 & 81.6 & 39.0 & 70.7 & 51.9 & 60.2 & 57.33 \\
\cmidrule(lr){1-9}
\rowoursalt \method{} (HB) & 16 & 49.7 & 86.8 & 47.1 & 73.7 & 57.5 & \textbf{67.5} & \textbf{63.70} \\
\rowoursalt \method{} (Random) & 16 & \stk{48.9}{1.7} & \stk{\textbf{87.0}}{0.3} & \stk{\textbf{48.2}}{1.1} & \stk{73.1}{0.7} & \stk{\textbf{59.0}}{1.4} & \stk{65.4}{1.9} & \stk{63.58}{0.94} \\
\rowours \method{} (Consecutive) & 16 & \textbf{50.0} & \textbf{87.0} & 47.1 & \textbf{73.9} & 56.5 & 67.1 & 63.60 \\
\bottomrule
\end{tabularx}

\end{table*}

\subsection{The Role of Calibration Inputs}
Mixed calibration trades 0.73--2.88 points of text \textit{\rev{Avg7}} for gains of 1.90--38.76 points in MM Avg across the configurations in Table~\ref{tab:app-mm-calib}. To distinguish the effects of grouping from those of activation geometry, we obtain partitions and whitening statistics from text-only or mixed calibration independently. For HB at $G=16$, mixed statistics with the text partition achieve 62.69 MM Avg, whereas the mixed partition with text statistics achieves 44.98. Most of the multimodal improvement therefore comes from representing image-conditioned activations in the reconstruction metric.
\begin{table}[!tb]
\caption{Calibration for multimodal accuracy on Qwen3.5-9B. Text calibration uses 256 FineWeb sequences, and mixed calibration 128 FineWeb sequences and 128 image-text examples. (a) $\Delta$ is mixed minus text. (b) The partition and the whitening statistics come from text or mixed calibration independently. \rev{Bold marks the higher Text}/\rev{Mixed accuracy for each metric in} (\rev{a}), \rev{and the best result for each metric in} (\rev{b}).}
\label{tab:app-mm-calib}
\centering
\scriptsize
\setlength{\tabcolsep}{2pt}
\tablecolors
\begin{tabularx}{\columnwidth}{@{\hspace{3pt}}cYYYYYYY@{\hspace{3pt}}}
\multicolumn{8}{@{\hspace{3pt}}l}{(a) Text-only and mixed calibration, HB} \\
\toprule
\rowband & & \multicolumn{3}{c}{\textit{\rev{Avg7}}$\uparrow$} & \multicolumn{3}{c}{MM~Avg$\uparrow$} \\
\cmidrule(lr){3-5}\cmidrule(lr){6-8}
\rowband CR & $G$ & Text & Mixed & $\Delta$ & Text & Mixed & $\Delta$ \\
\midrule
0.2 & 8 & \textbf{72.31} & 71.03 & -1.28 & 52.46 & \textbf{66.94} & +14.48 \\
0.2 & 16 & \textbf{72.63} & 70.91 & -1.72 & 65.59 & \textbf{67.49} & +1.90 \\
0.4 & 8 & \textbf{68.88} & 66.00 & -2.88 & 26.34 & \textbf{63.61} & +37.27 \\
0.4 & 16 & \textbf{69.88} & 67.71 & -2.17 & 35.87 & \textbf{63.70} & +27.83 \\
0.5 & 8 & \textbf{60.01} & 59.28 & -0.73 & 25.38 & \textbf{57.85} & +32.47 \\
0.5 & 16 & \textbf{62.92} & 61.91 & -1.01 & 20.89 & \textbf{59.65} & +38.76 \\
\bottomrule
\end{tabularx}

\medskip
\tablecolors
\begin{tabularx}{\columnwidth}{@{\hspace{3pt}}lYYYY@{\hspace{3pt}}}
\multicolumn{5}{@{\hspace{3pt}}l}{(b) Source of partition and statistics, HB, CR $0.4$, $G=16$} \\
\toprule
\rowband Partition & Statistics & PPL$\downarrow$ & \textit{\rev{Avg7}}$\uparrow$ & MM~Avg$\uparrow$ \\
\midrule
Text & Text & 3.43 & 69.88 & 35.87 \\
Text & Mixed & 3.70 & 67.41 & 62.69 \\
Mixed & Text & \textbf{3.34} & \textbf{70.05} & 44.98 \\
Mixed & Mixed & 3.56 & 67.71 & \textbf{63.70} \\
\bottomrule
\end{tabularx}

\end{table}

\section{Storage, Quantization, and Deployment}
\label{app:system}
This appendix distinguishes nominal factor allocation, packed model-weight storage, and measured execution time. These quantities describe complementary aspects of deployment and are evaluated under the configurations specified for each study.

\subsection{Effective Storage with Sparse Positions}\label{app:storage}
\paragraph{Sparse-position metadata.}
The nominal compression ratio counts dictionary entries and nonzero coefficient values. To account for sparse positions, we store the dictionary in bf16 (16 bits per value), each nonzero coefficient in 14 bits, and one bit per dictionary atom and output column in a support bitmask. Following CoSpaDi~\citep{shopkhoev2025cospadi}, we retain the upper 14 bits of each bf16 coefficient. Realizing this storage budget requires packing those 14-bit values, not merely zeroing two bits in a 16-bit array. The compression ratio of this packed representation is
\begin{equation}
\begin{aligned}
\mathrm{CR}_{\mathrm{real}}
&=1-\frac{16d\natoms+14\nnz M+\natoms M}{16dM}\\
&=\mathrm{CR}+\frac{2\nnz-\natoms}{16d}.
\end{aligned}
\label{eq:real-compression}
\end{equation}
Under the $\kappa=2.5$ allocation, support overhead reduces nominal CR by at most 1.25 percentage points with this coefficient format. The ratio refers to compressed attention and MLP projections, not necessarily to the entire model, because embeddings, output heads, and any uncompressed multimodal components remain dense.

\paragraph{Precision and support overhead.}
The packed representation uses no more bits than the nominal value-count budget when $\natoms\le2\nnz$. On Llama at nominal CR 0.4, the realized projection-level CR is 0.391--0.396 for $G=2$ and $G=4$, 0.400 for $G=8$, and 0.402 for $G=16$. Rounding the dictionary to bf16 and coefficients to 14 bits changes \textit{\rev{Avg7}} and LAMBADA perplexity by 0.01 each in a paired HB $G=16$ run.

With 16-bit coefficient values, the support bitmask lowers the realized CR to approximately 0.338 at $G=16$. Table~\ref{tab:app-storage} gives this alternative accounting across nominal budgets. Recovering a realized CR of 0.40 requires nominal CR 0.4625, and the resulting consecutive configurations achieve 63.91/67.58 \textit{\rev{Avg7}} on Llama/Qwen, compared with 66.61/70.58 at nominal CR 0.40. Precision and support encoding must therefore be specified alongside nominal compression when comparing storage--quality trade-offs.

\begin{table}[!tb]
\caption{\rev{Realized} CR (\%) of the compressed projections \rev{including coefficient-support metadata}, for nominal CR 20, 40 and 50\% (bf16 values, one bit per atom and output column for the support). The \rev{reduction relative to} nominal CR is $\natoms/(16d)$, at most 6.25 points once the dictionary is full width. \rev{Bold marks the largest realized CR within each model}/\rev{nominal-CR column}.}
\label{tab:app-storage}
\centering
\footnotesize
\setlength{\tabcolsep}{2pt}
\tablecolors
\begin{tabularx}{\columnwidth}{@{\hspace{3pt}}cYYYYYY@{\hspace{3pt}}}
\toprule
\rowband & \multicolumn{3}{c}{Llama-3-8B} & \multicolumn{3}{c}{Qwen3.5-9B} \\
\cmidrule(lr){2-4}\cmidrule(lr){5-7}
\rowband $G$ & CR~20 & CR~40 & CR~50 & CR~20 & CR~40 & CR~50 \\
\midrule
1 & \textbf{15.7} & \textbf{36.4} & \textbf{47.0} & \textbf{15.4} & \textbf{36.4} & \textbf{47.0} \\
2 & 15.2 & 35.6 & 46.0 & 14.7 & 35.4 & 46.1 \\
4 & 14.4 & 34.8 & 45.2 & 14.2 & 34.6 & 45.1 \\
8 & 13.8 & 34.2 & 44.5 & 13.8 & 34.1 & 44.3 \\
16 & 13.8 & 33.8 & 44.1 & 13.8 & 33.8 & 44.1 \\
\bottomrule
\end{tabularx}

\end{table}

\subsection{Whole-Model Storage and Compression Cost}
Table~\ref{tab:app-system} includes dense embeddings, output heads, and other uncompressed components in whole-model storage. Consecutive $G=16$ reduces the packed size from 16.06 to 10.45 GB on Llama and from 17.91 to 12.34 GB on Qwen. These figures describe weight storage. Runtime memory additionally depends on activations, workspaces, and the inference implementation.

Consecutive group construction requires no similarity search, while factor fitting uses each projection's activation Gram matrix. Median compression time on two A100 40GB GPUs is 3.88 hours for Llama and 2.97 for Qwen, compared with 1.32 and 1.11 for GeoPair.
\begin{table}[!tb]
\caption{\textbf{Whole-model storage and compression time.} GeoGroup uses consecutive $G=16$ at nominal CR $0.4$. Packed storage includes a bf16 dictionary, 14-bit coefficient values, a support bitmask, and uncompressed model components. Reduction is relative to dense bf16 storage. Time is the median compression wall time on two A100 40GB GPUs. \rev{Bold marks the smaller weight-storage value for each model}.}
\label{tab:app-system}
\centering\footnotesize
\renewcommand{\arraystretch}{1.10}
\setlength{\tabcolsep}{2pt}
\tablecolors
\begin{tabularx}{\columnwidth}{@{\hspace{3pt}}lYYYY@{\hspace{3pt}}}
\toprule
\rowband & Dense & GeoGroup & Reduction & Time\\
\rowband Model & (GB) & (GB) & (\%) & (h)\\
\midrule
Llama-3-8B & 16.06 & \textbf{10.45} & 35.0 & 3.88\\
Qwen3.5-9B & 17.91 & \textbf{12.34} & 31.1 & 2.97\\
\bottomrule
\end{tabularx}
\end{table}

\subsection{Compatibility with Factor Quantization}
\label{app:large}
Table~\ref{tab:app-large} expands the Qwen3.8-27B quantization results to individual tasks. Attention and MLP projections use nominal CR 0.4, with consecutive groups of eight for down projections and sixteen for the others. Factorization takes 9.9 hours on two GPUs. The 4-bit/4-bit and 8-bit/6-bit dictionary/coefficient configurations achieve 73.03 and 72.67 \textit{\rev{Avg7}}, respectively, compared with 74.37 for dense. These results demonstrate compatibility with reduced factor precision.
\begin{table*}[!t]
\caption{Qwen3.8-27B at CR $0.4$ (accuracy \%, LAMBADA PPL). The compressed configurations use the indicated dictionary and coefficient precisions. Consecutive groups contain 16 projections, except $^\dagger$down projections, which use groups of 8. Bold marks the highest accuracy and lowest perplexity among compressed configurations.}
\label{tab:app-large}
\centering
\footnotesize
\setlength{\tabcolsep}{2pt}
\tablecolors
\begin{tabularx}{\textwidth}{@{\hspace{3pt}}lYYYYYYYYYY@{\hspace{3pt}}}
\toprule
\rowband Method & $G$ & ARC-C & ARC-E & HellaS. & MMLU & PIQA & RACE & SciQ & \textit{Avg7}$\uparrow$ & PPL$\downarrow$ \\
\midrule
\dense{Dense} & \dense{--} & \dense{58.6} & \dense{73.1} & \dense{82.8} & \dense{83.5} & \dense{81.7} & \dense{47.6} & \dense{93.3} & \dense{74.37} & \dense{3.52} \\
\midrule
\method{}, $\mathbf{D}$ 8-bit, coefficients 6-bit & 16$^\dagger$ & 55.5 & 76.3 & \textbf{77.8} & \textbf{77.7} & \textbf{80.8} & 44.3 & 96.2 & 72.67 & \textbf{2.94} \\
\method{}, $\mathbf{D}$ 4-bit, coefficients 4-bit & 16$^\dagger$ & \textbf{56.6} & \textbf{81.9} & 76.3 & 74.8 & 79.7 & \textbf{45.5} & \textbf{96.5} & \textbf{73.03} & 3.09 \\
\bottomrule
\end{tabularx}

\end{table*}

\subsection{Selective Gate/Up Decode Performance}
\label{app:decode}
Main Table~\ref{tab:decode-summary} \rev{covers Qwen3.5-9B} and \rev{Qwen3.8-27B}. \rev{Supplementary} Table~\ref{tab:decode-gate} \rev{gives} gate/up component measurements for Qwen3.5-9B at batch size one. \rev{The compressed Qwen3.5-9B} configurations use MACKO. \rev{Each configuration is measured over} ten runs. The dense concatenated gate/up control separates the effect of concatenation from the additional effect of compression, and its weight storage matches the ordinary dense model.

Consecutive $G=16$ reduces full-model latency to 16.353 and 15.932 ms/token at CR 0.5 and 0.6. Pairwise compression \rev{achieves} lower latency and \rev{weight} storage in these measurements. Gate/up-only acceleration is reported separately from full-model acceleration.

CR denotes the nominal experimental setting rather than the measured reduction in bytes. For example, consecutive $G=16$ at CR 0.6 reduces gate/up weights from 6.44 to 3.20 GB and whole-model weights from 17.91 to 14.66 GB. Weight sizes in GB do not denote peak runtime memory. \rev{For Qwen3.5-9B}, \rev{values} after $\pm$ are standard deviations over runs, and $\pm0.000$ reflects rounding.

\paragraph{\rev{Qwen3.5-9B measurement} protocol.}
\rev{The Qwen3.5-9B} timings use one NVIDIA A100-SXM4-40GB GPU with FP16 weights. Each run prefills a 256-token prompt and greedily decodes 128 tokens, and every decode step of the whole network is replayed as one CUDA graph with a static cache. A run reports the mean time per token over the last 127 steps after an untimed verification pass. Every configuration uses ten runs. \rev{Gate}/up \rev{latency} is \rev{the difference between} step \rev{times with and} without the gate/up projections. Weight sizes exclude the vision encoder and the multi-token-prediction head.
\begin{table*}[!t]
\caption{Gate/up component timing on Qwen3.5-9B, batch size 1, for the same configurations as main Table~\ref{tab:decode-summary}. \rev{Latency and weight storage cover} gate/\rev{up projections only}. \rev{Storage is expressed in} GB. Times are the mean $\pm$ standard deviation over ten runs. Dense Cat is the dense control with concatenated gate/up projections, and the compressed configurations run with MACKO. Parentheses give reductions from dense weights. \rev{Bold marks the lowest latency and storage and the highest speedup among compressed configurations at each CR}.}
\label{tab:decode-gate}
\centering\small
\tablecolors
\begin{tabular}{lc@{\hspace{10pt}}ccccc}
\toprule
\rowband Model & $G$ & CR & ms/token & vs Dense & vs Cat & Weights (GB)\\
\midrule
Dense & -- & -- & $5.407 \pm 0.012$ & $1.00\times$ & -- & 6.44\\
Dense Cat & -- & -- & $5.103 \pm 0.018$ & $1.06\times$ & $1.00\times$ & 6.44\\
\rowgeopair GeoPair & 2 & 0.5 & $\mathbf{3.492} \pm 0.003$ & $\mathbf{1.55}\times$ & $\mathbf{1.46}\times$ & \textbf{3.80} ($-41\%$)\\
\rowgeopair GeoPair & 2 & 0.6 & $\mathbf{3.055} \pm 0.003$ & $\mathbf{1.77}\times$ & $\mathbf{1.67}\times$ & \textbf{3.04} ($-53\%$)\\
\rowours Consecutive & 8 & 0.5 & $4.183 \pm 0.001$ & $1.29\times$ & $1.22\times$ & 3.96 ($-38\%$)\\
\rowours Consecutive & 8 & 0.6 & $3.536 \pm 0.003$ & $1.53\times$ & $1.44\times$ & 3.17 ($-51\%$)\\
\rowours Consecutive & 16 & 0.5 & $4.291 \pm 0.004$ & $1.26\times$ & $1.19\times$ & 3.99 ($-38\%$)\\
\rowours Consecutive & 16 & 0.6 & $3.860 \pm 0.003$ & $1.40\times$ & $1.32\times$ & 3.20 ($-50\%$)\\
\bottomrule
\end{tabular}
\end{table*}

\paragraph{\rev{Qwen3.8-27B decode performance}.}
\rev{Table}~\ref{tab:decode-qwen27} \rev{reports full-model and component measurements at CR 0.6}. \rev{The ordinary dense and concatenated controls take 49.168 and 48.107 ms}/\rev{token}, \rev{respectively}. \rev{GeoPair} $\revmath{G}=\revmath{2}$ \rev{reaches 40.830 ms}/\rev{token} ($\revmath{1.20}\times$ \rev{versus dense}), \rev{while consecutive} $\revmath{G}=\revmath{16}$ \rev{reaches 43.571} ($\revmath{1.13}\times$). \rev{For gate}/\rev{up alone}, \rev{the corresponding speedups are} $\revmath{1.86}\times$ \rev{and} $\revmath{1.45}\times$. \rev{Gate}/\rev{up weight storage falls by 53}\% \rev{and 50}\%. \rev{CR 0.6 applies to gate}/\rev{up projections}, \rev{whereas the full-model storage reductions are 22}\% \rev{and 21}\%.

\rev{Measurements use batch size one and report mean} $\pm$ \rev{standard deviation over ten runs}. \rev{Accuracy and latency are evaluated on the same checkpoints} (\rev{Table}~\ref{tab:selective-quality}). \rev{Weight storage is expressed in decimal GB}, \rev{with parenthesized reductions relative to dense}.
\begin{table*}[!t]
\caption{\rev{Qwen3.8-27B with selective gate}/\rev{up compression and MACKO}, \rev{CR 0.6}. \rev{Cat denotes dense concatenated gate}/\rev{up}. \rev{Latency is reported as mean} $\pm$ \rev{SD over ten runs at batch size one}. \rev{Weights are in decimal GB}, \rev{with parenthesized reductions relative to dense}. \rev{Weight storage excludes runtime memory overhead}. \rev{Bold marks the best compressed result within each measurement scope}.}
\label{tab:decode-qwen27}
\centering\small\tablecolors
\begin{tabular}{lcrrrrr}
\toprule
\rowband \rev{Model} & $\revmath{G}$ & \rev{ms}/\rev{token} & \rev{tok}/\rev{s} & \rev{vs Dense} & \rev{vs Cat} & \rev{Weights} (\rev{GB})\\
\midrule
\multicolumn{7}{l}{\textbf{\rev{Full-model decode}}}\\
\rev{Dense} & -- & $\revmath{49.168} \pm \revmath{0.004}$ & \rev{20.3} & $\revmath{1.00}\times$ & -- & \rev{53.79}\\
\rev{Dense Cat} & -- & $\revmath{48.107} \pm \revmath{0.016}$ & \rev{20.8} & $\revmath{1.02}\times$ & $\revmath{1.00}\times$ & \rev{53.79}\\
\rowgeopair \rev{GeoPair} & \rev{2} & $\revmath{\mathbf{40.830}} \pm \revmath{0.001}$ & \textbf{\rev{24.5}} & $\revmath{\mathbf{1.20}}\times$ & $\revmath{\mathbf{1.18}}\times$ & \textbf{\rev{41.78}} ($-\revmath{22}\%$)\\
\rowours \rev{Consecutive} & \rev{16} & $\revmath{43.571} \pm \revmath{0.002}$ & \rev{23.0} & $\revmath{1.13}\times$ & $\revmath{1.10}\times$ & \rev{42.29} ($-\revmath{21}\%$)\\
\midrule
\multicolumn{7}{l}{\textbf{\rev{Gate}/\rev{up component only}}}\\
\rev{Dense} & -- & $\revmath{18.030} \pm \revmath{0.030}$ & -- & $\revmath{1.00}\times$ & -- & \rev{22.82}\\
\rev{Dense Cat} & -- & $\revmath{16.979} \pm \revmath{0.018}$ & -- & $\revmath{1.06}\times$ & $\revmath{1.00}\times$ & \rev{22.82}\\
\rowgeopair \rev{GeoPair} & \rev{2} & $\revmath{\mathbf{9.713}} \pm \revmath{0.006}$ & -- & $\revmath{\mathbf{1.86}}\times$ & $\revmath{\mathbf{1.75}}\times$ & \textbf{\rev{10.80}} ($-\revmath{53}\%$)\\
\rowours \rev{Consecutive} & \rev{16} & $\revmath{12.450} \pm \revmath{0.007}$ & -- & $\revmath{1.45}\times$ & $\revmath{1.36}\times$ & \rev{11.32} ($-\revmath{50}\%$)\\
\bottomrule
\end{tabular}
\end{table*}

\end{document}